\documentclass[10pt,journal,twoside,onecolumn]{IEEEtran}

\usepackage{changepage}

\usepackage{psfrag,epsfig,graphics}
\usepackage{amsmath,amssymb,multirow}

\usepackage{balance}

\usepackage{graphicx}
\usepackage{caption}
\usepackage[position=b]{subcaption}

\usepackage[noadjust]{cite}
\usepackage{multirow}
\usepackage[lined,linesnumbered,ruled]{algorithm2e}

\usepackage{color}  % To include comments in color

\newcommand{\cb}[1]{{\boldsymbol{#1}}}
\newcommand{\cp}[1]{\ifmmode {\mathcal{#1}}\else ${\mathcal{#1}}$\fi}
\newcommand{\balpha}{\boldsymbol{\alpha}}
\newcommand{\bbeta}{\boldsymbol{\beta}}
\newcommand{\bgamma}{\boldsymbol{\gamma}}

\newcommand{\bI}{\boldsymbol{I}}
\newcommand{\bK}{\boldsymbol{K}}

\newcommand{\bc}{\boldsymbol{c}}

\newcommand{\bm}{\boldsymbol{m}}
\newcommand{\bn}{\boldsymbol{n}}

\newcommand{\be}{\boldsymbol{e}}

\newcommand{\br}{\boldsymbol{r}}

\newcommand{\psh}[2]{\langle{#1},{#2}\rangle_{\cp{H}}}

\newcommand{\mr}[1]{{\bm_{\lambda_{#1}}}}

\newcommand{\MM}{{\boldsymbol{M}}}

\newcommand{\KK}{\boldsymbol{K}}

\newcommand{\adjM}{{\boldsymbol{B}}}

\begin{document}

%\title{\Huge{Redundancy reduction in RKHS for nonlinear unmixing of hyperspectral images}}

%\title{\Huge{Fast sparse regression in RKHS for nonlinear unmixing of hyperspectral images}}
%\title{\Huge{Coherence-based band selection in RKHS for fast nonlinear unmixing of hyperspectral images}}
\title{\Huge{Technical Report:\\ Band selection for nonlinear unmixing of hyperspectral images as a maximal clique problem}}
\author{ Tales Imbiriba$^{(1)}$, \emph{Student Member, IEEE}, Jos{\'e} Carlos Moreira Bermudez$^{(1)}$, \emph{Senior Member, IEEE},\\ C{\'e}dric Richard$^{(2)}$, \emph{Senior Member, IEEE}
%\\ \vspace{0.5cm}
% \small{$^{(1)}$ Universidade Federal de Santa Catarina, 88040-900\\
% Departamento de Engenharia El{\'e}trica, Florian{\'o}polis, Santa Catarina - Brazil. \\
% tel.: +55.48.37.21.77.19 \hspace{0.5cm} \hspace{0.5cm} fax.:
% +55.48.37.21.77.19 \\
% talesim@gmail.com \hspace{0.5cm} j.bermudez@ieee.org}
% \vspace{0.3cm}\\
% \small{\linespread{0.2}$^{(2)}$ Universit{\'e} de Nice Sophia-Antipolis, UMR CNRS 7293, Observatoire de la C{\^{o}}te d'Azur \\
% Laboratoire Lagrange, Parc Valrose, 06102 Nice cedex 2 - France \\
% tel.: +33.4.92.07.63.94 \hspace{0.5cm} \hspace{0.5cm} fax.:
% +33.4.92.07.63.21 \\
% cedric.richard@unice.fr}
% \vspace{0.3cm}\\
% \small{\linespread{0.2}$^{(3)}$ Universit{\'e} de Toulouse\\
% IRIT-ENSEEIHT, Toulouse 31071 - France \\
% tel.: +.... \hspace{0.5cm} \hspace{0.5cm} fax.:
% +.... \\
% jean-yves.tourneret@enseeiht.fr}
% \vspace{0.3cm}\\
% 
% \thanks{This work was partly supported by CNPq under grants Nos 305377/2009-4, 400566/2013-3 and 141094/2012-5, and by the Agence Nationale pour la Recherche, France, (Hypanema project, ANR-12- BS03-003), and by ANR-11-LABX-0040-CIMI within the program ANR-11-IDEX-0002-02.}

\thanks{T. Imbiriba and J.-C. M. Bermudez are with the Department of Electrical Engineering, Federal University of Santa Catarina at Florian{\'o}polis, SC, 88040-900, Brazil. C. Richard is with the University of Nice Sophia-Antipolis, Nice 06108, France (e-mail: cedric.richard@unice.fr), Lagrange Laboratory (CNRS, OCA), in collaboration with Morpheme team (INRIA Sophia-Antipolis). The work of J.-C. M. Bermudez was partly supported by Conselho Nacional de Desenvolvimento Científico e Tecnológico (CNPq) grants 305377/2009-4, 400566/2013-3 and 141094/2012-5. The work of C. Richard was partly supported by ANR grants ANR-12- BS03-003 (Hypanema), by the CNRS Imag’in project under grant 2015OPTIMISME, by the BNPSI ANR Project no ANR-13- BS-03-0006-01.}

}

\maketitle
% \vspace{1cm}
\begin{abstract}
Kernel-based nonlinear mixing models have been applied to unmix spectral information of hyperspectral images when the type of mixing occurring in the scene is too complex or unknown. Such methods, however, usually require the inversion of matrices of sizes equal to the number of spectral bands. Reducing the computational load of these methods remains a challenge in large scale applications. This paper proposes a centralized method for band selection (BS) in the reproducing kernel Hilbert space (RKHS). It is based upon the coherence criterion, which sets the largest value allowed for correlations between the basis kernel functions characterizing the unmixing model. We show that the proposed BS approach is equivalent to solving a maximum clique problem (MCP), that is, searching for the biggest complete subgraph in a graph. Furthermore, we devise a strategy for selecting the coherence threshold and the Gaussian kernel bandwidth using coherence bounds for linearly independent bases. Simulation results illustrate the efficiency of the proposed method.
\end{abstract}


\section{Introduction}

The unmixing of spectral information acquired by hyperspectral sensors is at the core of many remote sensing applications such as land use analysis, mineral detection, environment monitoring and field surveillance~\cite{Bioucas-Dias-2013-ID307, Manolakis:2002p5224}. Such information is typically mixed at the pixel level due to the low resolution of hyperspectral devices or because distinct materials are combined into a homogeneous mixture~\cite{Keshava:2002p5667}. The observed reflectances then result from mixtures of several pure material signatures present in the scene, called endmembers. Considering that the endmembers have been identified, hyperspectral unmixing (HU) refers to estimating the proportional contribution of each endmember to each pixel in a scene. 

The linear mixture model is widely used to identify and quantify pure components in remotely sensed images due to its simple physical interpretation. Though the linear model leads to simple unmixing algorithms and facilitates implementation, there are many situations to which it is not applicable. These include scenes where there is complex radiation scattering among several endmembers, as may happen in some vegetation areas~\cite{Ray1996}. In such situations, nonlinear mixing models must be considered~\cite{heylen2014review,Dobigeon-2014-ID322}. Several nonlinear mixing models have been proposed in the literature. A review of the existing models can be found in~\cite{heylen2014review}. The complexity of the mixture mechanisms that may be present in a real scene has led to the consideration of flexible nonlinear mixing models that can model generic nonlinear functions. Kernel methods provide a non-parametric representation of functional spaces, and can model nonlinear mixings of arbitrary characteristics~\cite{Wu2010,Li2012blind,heylen2014review,Dobigeon-2014-ID322, Chen-2013-ID321, chen2013nonlinear}.

Kernel-based methods are efficient machine learning techniques~\cite{Vapnik1995, smola2004tutorial, Suykens2002} that consist mainly of linear algorithms operating in high dimensional reproducing kernel Hilbert spaces (RKHS), into which the data have been mapped using kernel functions~\cite{Vapnik1995}. Working in such high dimensional feature spaces is possible due to the so-called \emph{kernel trick}, which allows the computation of inner products in the feature space through a kernel function in the input space~\cite{Aronszajn1950}. A limitation of kernel methods for HU is that they usually require the inversion of matrices whose dimensions equal the number of spectral bands. Thus, reducing their computational cost remains a challenge for their use in large-scale applications.

A possible way of reducing this cost is to perform band selection (BS) prior to unmixing~\cite{chang2014hyperspectral}.  Though BS has been actively employed in classification of spectral patterns~\cite{du2008similarity, estevez2009normalized, martinez2007clustering,Feng-2014-ID352, Feng-2015-ID338}, subspace projection techniques~\cite{BioucasDias:2008gc,Nascimento2005,BioucasDias:2011fi} tend to be preferred over BS~\cite{Chang-2006-ID353, chang2014progressive} for reducing the complexity of linear unmixing processes. This is mainly because high-dimensional data are confined to a low-dimensional simplex in linearly-mixed images with only a few endmembers~\cite{Keshava:2002p5667}. However, the simplex property is not preserved in the presence of nonlinearly-mixed pixels~\cite{Dobigeon-2014-ID322}, rendering projection techniques less attractive. Nevertheless, BS is also a challenging problem for nonlinear unmixing since the selection procedure should ideally match the characteristics of the unmixing model. Thus, BS methods developed for linear mixed pixels cannot be directly applied to the nonlinear case. 

In a previous work~\cite{Imbiriba2015}, we proposed a BS method based on the kernel $k$-means algorithm to identify clusters of spectral bands in the corresponding RKHS. The cluster prototypes are then the selected bands. This method reduces significantly the computation time required for nonlinear unmixing without compromising the accuracy of abundance estimation. In this approach, however, each band is selected based on its distance to the others in the RKHS, and not as a function of the resulting accuracy of the unmixing procedure. In addition, it requires to set the final number of bands \emph{a priori}. Hence, some cluster prototypes can be close to others and degrade problem conditioning if this parameter is overestimated.

 In~\cite{Richard2009}, the authors proposed a low-complexity coherence-based greedy approach for controlling the size of kernel models for online system identification. As the coherence criterion makes the needed bridge between the number of basis kernel functions in the unmixing model and an upper bound on the reconstruction error, such approach could may also be applied to BS in RKHS.  However, its greedy nature which is appropriate for online settings would lose efficiency otherwise.

In this paper we introduce a new coherence-based method for BS in the RKHS. The coherence criterion is used to set the largest correlation between the basis kernel functions included in the unmixing model. We show that this BS approach is equivalent to search for a maximum clique in a graph, that is, the largest complete subgraph in this graph. Starting from a tentative dictionary cardinality, the proposed method determines both the dictionary size and its elements in order to satisfy the required coherence criterion. Using the  maxCQL algorithm~\cite{li2010efficient} to solve the maximum clique problem, the new method results in dictionaries of kernel functions, and thus spectral bands, that are less coherent than those obtained using kernel $k$-means initialized with dictionaries of the same size. 

% The maximum clique problem (MCP) has been a topic of great interest in the past decades~\cite{pardalos1994maximum, ostergaard2002fast} and still attracts considerable attention~\cite{wu2015review, maslov2014speeding, li2013combining}. MCP is NP-complete~\cite{karp1972reducibility}, and finding exact solutions for large graphs is not guaranteed. Recently, several methods have however been proposed in order to produce fast and near-optimal solutions under a variety of conditions including large dense graphs~\cite{li2010efficient, li2013combining, maslov2014speeding}. 

This paper is organized as follows. First, we review nonlinear unmixing models usually considered for HU. Then, we introduce kernel-based models and the associated estimation framework. Next, we consider the problem of BS in RKHS. We introduce kernel $k$-means strategy and our new algorithm based on maximum clique search. We provide promising simulation results to illustrate the performance of the proposed method using both synthetic and real images. Finally, we present some concluding remarks.

\section{Hyperspectral Images and unmixing}
\label{sec:HI}

Observed pixels in HIs are usually modeled as a function, possibly nonlinear, of the endmembers and an additive noise that accounts for the measurement noise plus a modeling error, namely,
\begin{equation}
       \label{eq:model}
        \br = {\cb\psi}(\MM) + \cb{n}
\end{equation}
where $\br =[r_1,\ldots, r_L]^\top$ is a vector of observed reflectances in $L$ spectral bands, $\MM = [\bm_1,\ldots,\bm_R]$ is the $L\times R$ matrix of R endmembers, whose $i$-th column $\bm_i$ corresponds to an endmember, $\bn$ is a white Gaussian noise (WGN) vector, and function $\cb{\psi}$ represents an unknown mixing mechanism. Several models of the form \eqref{eq:model} were proposed in the literature, depending on the linearity or nonlinearity of $\cb{\psi}$, the nature of mixture, and other properties~\cite{Dobigeon-2014-ID322}.

\subsection{The linear mixing model}

The linear mixing model (LMM) considers only interactions of light rays with a single material, neglecting interactions between light and several materials~\cite{Keshava:2002p5667}. The LMM assumes that $\br$ is a convex combination of the endmembers, namely,
\begin{equation}
\begin{split}
	&\br = \MM \balpha + \bn\\
	&\text{subject to }\,\cb{1}^\top\balpha = 1 \text{ and } \balpha \succeq \cb{0}
 \label{eq:linForm}
\end{split}
\end{equation}
where $\balpha = [\alpha_1,\ldots,\alpha_{R}]^\top$ denotes the vector of abundances of each endmember in $\MM$,  and $\succeq$ is the entrywise $\geq$ operator. Being proportions, the entries of $\balpha$ cannot be negative and should sum to one. The observation $r_\ell$ in the $\ell$-th wavelength of~\eqref{eq:linForm} can be written as 
\begin{equation}
	r_\ell = \bm_{\lambda_\ell}^\top\balpha + n_\ell
	\label{eq:LMM_band}
\end{equation}
where $\bm_{\lambda_\ell}$ denotes the $\ell$-th row of $\MM$ written as a column vector.
In the noiseless case ($n_\ell = 0$), the sum-to-one and positivity constraints over $\balpha$ in~\eqref{eq:linForm} restrict the data to a simplex whose vertices are the endmembers.

\subsection{Nonlinear mixing models}

Several nonlinear models have been proposed to describe complex mixing mechanisms. See~\cite{Dobigeon-2014-ID322} and references therein. We now review two popular models that will be used later. 

The generalized bilinear model (GBM)~\cite{halimi2011} is defined as:
\begin{equation}
	\label{eq:GBM_orig}
	\begin{split}
		&\br = \MM\balpha + \sum_{i=1}^{R-1}\sum_{j=i+1}^{R}\delta_{ij}\,\alpha_i\alpha_j\,\bm_i\odot\bm_j + \bn \\
		&\text{subject to }\cb{1}^\top\balpha = 1 \text{ and } \balpha \succeq 0
	\end{split}
\end{equation}
where each parameter $\delta_{ij}\in[0,1]$ characterizes the interaction of endmembers $\bm_i$ and $\bm_j$, and $\odot$ denotes the Hadamard product. For simplicity, we shall consider a simplified version of this model where all the bilinear terms in \eqref{eq:GBM_orig} are weighted by a single parameter $\delta = \delta_{ij}$ for all $(i,j)$.

The post nonlinear mixing model (PNMM)~\cite{Jutten2003} is defined as follows:
\begin{equation}
	\br = \cb{g}(\MM\balpha) + \bn
	\label{eq:PostLinearModel}
\end{equation}
where $\cb{g}$ is a nonlinear function applied to the noiseless LMM. Thanks to function $\cb{g}$, the PNMM specifies a large family of nonlinear mixing models via a single expression. For instance, the PNMM considered in~\cite{chen2013nonlinear} is given by
\begin{equation}
	\label{eq:pnmm_chen}
	\br = (\MM\balpha)^\xi + \bn
\end{equation}
where $(\cb{v})^{\xi}$ denotes the exponentiation applied to each entry of $\cb{v}$. For $\xi=2$, \eqref{eq:pnmm_chen} is a bilinear model closely related to the GBM but without a linear term.  The PNMM has been explored with different forms for $\cb{g}$~\cite{Altmann-2013-ID308,altmann2011:icassp}. 

The GBM and the PNMM models essentially describe situations where the light interacts first with an endmember, and then with a second one, before being captured by the hyperspectral sensor. Other nonlinear models can be considered depending on the characteristics of the scene~\cite{fan2009, Nascimento2009, Jutten2003, halimi2011,HapkeBook1993,Borel:1994tp,Somers:2009p6577,Ray:1996tp,Broadwater2009}. More importantly, information about these characteristics is usually missing, and it makes sense to consider nonparametric models that do not rely on strong assumptions.

\section{LS-SVR for hyperspectral unmixing}

Kernel-based methods consist of mapping observations from the original input space into a feature space by means of a nonlinear function. Nonlinear regression problems can be addressed in an efficient way in this new space as they are converted to a linear problem. We shall now review the main definitions related to RKHS~\cite{Kimeldorf1971,kreyszig1989introductory,scholkopf2001generalized,smola2004tutorial}.

\subsection{Mercer kernels and RKHS}

The theory of positive definite kernels emerged from the study of positive definite integral operators~\cite{Mercer1909}, and was further generalized for the study of positive definite matrices~\cite{Moore1916}. It was established that, to every positive definite function
\begin{equation}
	\kappa:\cp{M}\times \cp{M}\rightarrow\mathbb{R}
\end{equation}
defined over a non-empty compact $\cp{M}\subset\mathbb{R}^d$, there corresponds one and only one family of real-valued functions on $\cp{M}$ that defines a Hilbert space $\cp{H}$ endowed with an unique inner product $\psh{\cdot}{\cdot}$ and the associated norm $\|\!\cdot\!\|_{\mathcal{H}}$, and admitting $\kappa$ as a reproducing kernel~\cite{Aronszajn1950}. This means that $\kappa(\cdot,\cb{m})\in\cp{H}$ for all $\cb{m}\in \cp{M}$, and has the reproducing property defined as:
\begin{equation}
	\label{eq:reproducingProp}
	\psi(\cb{m}) = \psh{\psi}{\kappa(\cdot, \cb{m})}
\end{equation}
 for all $\psi\in\cp{H}$ and $\cb{m}\in\cp{M}$. Replacing $\psi$ by $\kappa(\cdot, \cb{m}')$ in~\eqref{eq:reproducingProp} leads to:
\begin{equation}
	\label{eq:reproducingKer}
	\kappa(\cb{m},\cb{m}') = \psh{\kappa(\cdot, \cb{m})}{\kappa(\cdot, \cb{m}')}
\end{equation}
for all $\cb{m},\cb{m}'\in\cp{M}$. Equation~\eqref{eq:reproducingKer} is the origin of the now generic denomination reproducing kernel to refer to $\kappa$. Note that $\cp{H}$ can be restricted to the span of $\{\kappa(\cdot,\cb{m}):\cb{m}\in\cp{M}\}$ because, according to the reproducing property \eqref{eq:reproducingProp}, nothing outside this set affects $\psi$ evaluated at any point of $\cp{M}$. Let us denote by $\varphi$ the map from $\cp{M}$ to $\cp{H}$ that assigns $\kappa(\cdot,\bm)$ to~$\bm$. Relation~\eqref{eq:reproducingKer} implies that $\kappa(\cb{m},\cb{m}')=\psh{\varphi(\cb{m})}{\varphi(\cb{m}')}$. This means that the kernel $\kappa$ evaluates the inner product of any pair of elements of $\cp{M}$ mapped into $\cp{H}$ without any explicit knowledge of $\varphi$ or $\cp{H}$. This principle is called the kernel trick.

Several kernel functions have been considered in a variety of applications during the past two decades~\cite{ScholkopfBook:2001}. Among the most frequently used kernels, we highlight the Gaussian kernel:
\begin{equation}
	\label{eq:gaussian-kernel}
	\kappa(\bm,\bm') = \exp\left(-\frac{\|\bm-\bm'\|^2}{2\sigma^2}\right)
\end{equation}
where $\sigma$ is the kernel bandwidth.

\subsection{LS-SVR: least squares support vector regression}

This section describes the use of a state-of-the-art kernel method for nonlinear unmixing of hyperspectral data. Consider an observation $r_\ell$ at the $\ell$-th wavelength, that is, the $\ell$-th entry of $\br$, and the column vector $\mr{\ell}$ of the $R$ endmember signatures at the $\ell$-th wavelength, that is, the (transposed) $\ell$-th row of $\cb{M}$. By analogy with the LMM~\eqref{eq:LMM_band}, we write: 
\begin{equation}
	r_\ell = \psi(\mr{\ell}) + n_\ell
\end{equation}
with $\psi$ a real-valued function in a RKHS $\cp{H}$ that characterizes the nonlinear interactions between the endmembers, and $n_\ell$ an additive noise at the $\ell$-th band. In order to estimate $\psi$ in the least squares sense, we can formulate the following convex optimization problem, also called LS-SVR~\cite{Suykens2002}:
\begin{equation}
	\begin{split} 
      		\label{eq:App_ls-svm}
      		&\mathop{\min}_{\psi\in\cp{H}}\,\, \frac{1}{2}\|\psi\|_\cp{H}^2+\frac{1}{2\mu}\sum_{\ell=1}^L e_\ell^2 \\
      		& \text{such that} \quad e_\ell = r_\ell- \psi(\mr{\ell}), \quad \ell=1,\dots,L.
      	\end{split}
\end{equation}
Consider the Lagrangian function
\begin{equation}
	\label{eq:AppLagrangean1}
	\cp{L}(\psi, \be, \bbeta) 
	= \frac{1}{2}\|\psi\|_\cp{H}^2+\frac{1}{2\mu}\sum_{\ell=1}^L e_\ell^2 - \sum_{\ell=1}^L \beta_\ell\,(e_\ell - r_\ell + \psi(\mr{\ell})).
\end{equation}
where $\bbeta=[\beta_1,\ldots,\beta_L]^\top$ is the vector of Lagrange multipliers. Using the directional derivative with respect to $\psi$~\cite{kadri2009}, the conditions for optimality with respect to the primal variables $\psi$ and $e_\ell$ are given by
\begin{align}
		\psi^* &= \sum_{\ell=1}^L\beta_\ell\kappa(.,\mr{\ell}) 		\label{appEq:gradFuncL.1}\\
		e_\ell^* &= \mu\beta_\ell							\label{appEq:gradFuncL.2}
\end{align}
Substituting \eqref{appEq:gradFuncL.1} and \eqref{appEq:gradFuncL.2} in \eqref{eq:AppLagrangean1}, we obtain the following function to be maximized with respect to $\bbeta$:
\begin{equation}
	\mathcal{L}(\psi^*,\be^*, \bbeta)=-\frac{1}{2}\,\bbeta^\top\!\left(\bK+\mu\cb{I}\right)\bbeta+\bbeta^\top\br,
\end{equation}
where $\KK$ is the Gram matrix whose $(i,j)$-th entry is defined by $\kappa(\mr{i},\mr{j})$. Now we can state the following dual problem:
\begin{equation}
	\label{eq:Appcost.nonlinear.init}
      	\bbeta^* = \mathop{\arg\max}_{\bbeta} -\frac{1}{2}\,\bbeta^\top\!\left(\bK+\mu\cb{I}\right)\bbeta+\bbeta^\top\br.
\end{equation}
Its solution is obtained by solving the linear system:
\begin{equation}
	\label{eq:dualSystem}	
      		\left(
		\begin{array}{c|c}
		   \multirow{2}{*}{$-\cb{I}$} & \multirow{2}{*}{$\cb{K} + \mu\cb{I}$}\\
		   %\multirow{2}{*}{-I} & \multirow{2}{*}{b}\\
		   &
		\end{array}
		\right)
		\left(
		\begin{array}{c}
			\br \\ \hline
 			\bbeta
		\end{array}\right)			
		=
		\cb{0}.
\end{equation}
Although the formulation \eqref{eq:App_ls-svm}--\eqref{eq:Appcost.nonlinear.init} allows one to address an estimation problem in $\cp{H}$ by solving the linear system~\eqref{eq:dualSystem}, this approach is computationally demanding since it involves the inversion of $L \times L$ matrices. This issue is critical, as modern hyperspectral image sensors employ hundreds of contiguous bands with an ever increasing spatial resolution. Hence, it is of major interest to consider band selection techniques that lead to significant computational cost reduction without noticeable quality loss. Considering~\eqref{appEq:gradFuncL.1}, a possible strategy is to focus on a reduced-order model of the form:
\begin{equation}
	\psi = \sum_{j\in\cp{I}_D}\beta_j \kappa(.,\mr{j})
\end{equation}
where $\cp{I}_D\subset\{1,\ldots,L\}$ is an $M$-element ($M<L$) subset of indexes. We shall call $\cp{D}=\{\kappa(.,\mr{j})\}_{j\in\cp{I}_D}$ the dictionary.              

%%%%%%%

\section{Band Selection}
\label{sec:BS}

BS has been an active topic of research for classification of spectral patterns, see~\cite{du2008similarity, estevez2009normalized, martinez2007clustering,Feng-2014-ID352, Feng-2015-ID338} and references therein. Subspace projection techniques~\cite{BioucasDias:2008gc,Nascimento2005,BioucasDias:2011fi} tend, however, to be preferred over BS~\cite{Chang-2006-ID353, chang2014progressive} for reducing the complexity of linear unmixing processes. They use the property that high-dimensional hyperspectral data are confined to a low-dimensional simplex in linearly-mixed images with only a few endmembers~\cite{Keshava:2002p5667}. This assumption becomes invalid when nonlinear mixing phenomena are involved. Recently, in a preliminary work~\cite{Imbiriba2015}, we introduced a BS strategy method that employs the kernel $k$-means algorithm to identify clusters of spectral bands in the RKHS where nonlinear unmixing is performed. The HU results obtained were encouraging.  One drawback of the approach in~\cite{Imbiriba2015} is the need for an arbitrary choice of the order of the nonlinear model (the dimension of the dictionary). Given the order, band selection is performed based on the distances among different bands in the RKHS. Hence, the optimality of the solution is not driven by any direct measure of modeling accuracy. In this section, we briefly review the kernel $k$-means approach. Then we introduce a new strategy based on the so-called coherence criterion~\cite{Richard2009} and maximum clique search in a graph. Although these two approaches are connected, they differ in their formulation and in the characteristics of the sets of bands they select.

\subsection{Kernel $k$-means for band selection}

Kernel $k$-means (KKM) is a direct extension of the $k$-means clustering algorithm~\cite{Tzortzis-2009-ID339}. It maps the input data $\mr{\ell}$ into a RKHS $\cp{H}$, and groups their images $\kappa(\cdot,\mr{\ell})$ into disjoint clusters $\cp{C}_1,\ldots,\cp{C}_M$ based on their relative distance in $\cp{H}$.  Since determining centroids in $\cp{H}$ is intractable, KKM calculates distances using the reproducing property~\eqref{eq:reproducingKer}.

Given a cluster $\cp{C}_k$ enclosing points $\{\kappa(\cdot,\mr{\ell})\}_{\ell\in \cp{C}_k}$, its centroid is defined as
\begin{equation}
	\nu_k = \frac{1}{N_k} \sum_{i\in\cp{C}_k}\kappa(\cdot, \mr{i})
\end{equation}
where $N_k$ is the number of points in $\cp{C}_k$. The squared distance of any point $\kappa(\cdot,\mr{\ell})$ to $\nu_k$ is computed as
\begin{equation}
	\begin{split}
 		\|\kappa(\cdot,\mr{\ell}) - \nu_k\|^2_{\cp H} &= \kappa(\mr{\ell},\mr{\ell})\\
 		&\quad - \frac{1}{N_k} \sum_{i\in \cp{C}_k}\kappa(\mr{\ell},\mr{i})\\
 		&\quad + \frac{1}{N_k^2}\sum_{i\in \cp{C}_k}\sum_{j\in \cp{C}_k}\kappa(\mr{i},\mr{j})
 	\end{split}
\end{equation}
and the clustering error to minimize is defined as
\begin{equation}
	E(\nu_1,\ldots,\nu_K) = \sum_{k=1}^{M}\sum_{\ell\in \cp{C}_k} \| \kappa(\cdot,\mr{\ell}) - \nu_k\|^2_{\cp H}.
\end{equation}
Each cluster $\cp{C}_k$ is then represented by the band $\ell_k$ corresponding to the closest point to its centroid $\nu_k$:
\begin{equation}
	\label{eq:bandSelection}
	\ell_k =  \mathop{\arg\min}_{\ell\in\cp{C}_k} \|\kappa(\cdot,\mr{\ell}) - \nu_k\|^2_{\cp{H}}.
\end{equation}
The global kernel $k$-means (GKKM) algorithm uses the principles described above for incremental clustering~\cite{Tzortzis-2009-ID339}. GKKM  does not suffer from poor convergence to local minima and produces near-optimal solutions that are robust to cluster initialization. A fast GKKM (FGKKM) version that performs a unique KKM run and greatly reduces the complexity of the algorithm can also be used. For more details on KKM for BS, the reader is invited to refer to~\cite{Imbiriba2015}.

\subsection{Coherence criterion for dictionary selection}

Coherence is a parameter of fundamental interest for characterizing dictionaries of atoms in linear sparse approximation problems~\cite{tropp2004greed}. It was first introduced as an heuristic quantity for Matching Pursuit in~\cite{mallat1993matching}. Formal studies followed  in~\cite{donoho2001uncertainty}, and were enriched for Basis Pursuit in~\cite{elad2002generalized,donoho2003optimally}. 

Consider a set of kernel functions $\{\kappa(\cdot,\mr{\ell})\}_{\ell=1,\ldots,M}$ in $\cp{H}$. The definition of coherence was extended to RKHS as~\cite{Richard2009}:
\begin{equation}
	\label{eq:coherencePar}
	\begin{split}
	\mu 	&= \max_{i\neq j} |\psh{\kappa(\cdot,\mr{i})}{\kappa(\cdot,\mr{j})}| \\
		&= \max_{i\neq j} |\kappa(\mr{i},\mr{j})|
	\end{split}
\end{equation}
where $\kappa$ is a unit-norm kernel. Otherwise, replace $\kappa(\cdot,\mr{i})$ with $\kappa(\cdot,\mr{i})/\sqrt{\kappa(\mr{i},\mr{i})}$ in~\eqref{eq:coherencePar}. Parameter $\mu$ is the largest absolute value of the off-diagonal entries in the Gram matrix. It reflects the largest cross correlation in the dictionary $\{\kappa(\cdot,\mr{\ell})\}_{\ell}$, and is equal to zero for every orthonormal basis. A dictionary is said to be incoherent when its coherence $\mu$ is small. Although its definition is rather simple, coherence possesses important properties~\cite{Richard2009}. In particular, it can be shown that the kernel functions in the dictionary $\cp{D}=\{\kappa(\cdot,\mr{\ell})\}_{\ell=1,\ldots,M}$ are linearly independent if $(M-1)\mu<1$.  This sufficient condition illustrates that the coherence \eqref{eq:coherencePar} provides valuable information on a dictionary at low computionnal cost. Other properties are discussed in~\cite{Richard2009}. 

Kernel-based dictionary learning methods usually consider approximate linear dependence conditions to evaluate whether a candidate kernel function $\kappa(\cdot,\mr{i})$ can be reasonably well represented by a combination of the kernel functions that are already in the dictionary $\cp{D}$. To avoid excessive computational complexity, a greedy dictionary learning method has been introduced in~\cite{Richard2009}. It consists of inserting the candidate $\kappa(\cdot,\mr{i})$ into the dictionary $\cp{D}$ provided its coherence is still below a given threshold $\mu_0$, namely,
\begin{equation}
	\mathop{\max}_{j\in \cp{I_D}}|\kappa(\mr{i},\mr{j})| \leq \mu_0
	\label{eq:coherenceThreshold}
\end{equation}
where $\mu_0$ is a parameter $[0,1[$ determining both the maximum coherence in $\cp{D}$ and its cardinality $|\cp{D}|$. Using coherence criterion for BS allows to explicitly limit the correlation of kernel functions in the dictionary. This contrasts with the kernel $k$-means strategy, which starts from a number of dictionary elements prescribed by the user without taking the coherence of kernel functions into consideration.

The coherence criterion~\eqref{eq:coherenceThreshold} was proposed within the context of parameter estimation from streaming data. The design of the dictionary follows a greedy strategy. The first kernel function is selected arbitrarily, and each new candidate kernel function is tested using \eqref{eq:coherenceThreshold} to determine if it deserves being included in the dictionary. This procedure is appropriate for online applications because of its minimal computational cost. However, alternatives should be sought which may lead to more effective solutions in batch mode applications.

\subsection{Band selection as a maximum clique problem}

\begin{figure}
	\centering
	\includegraphics[width=0.25\textwidth]{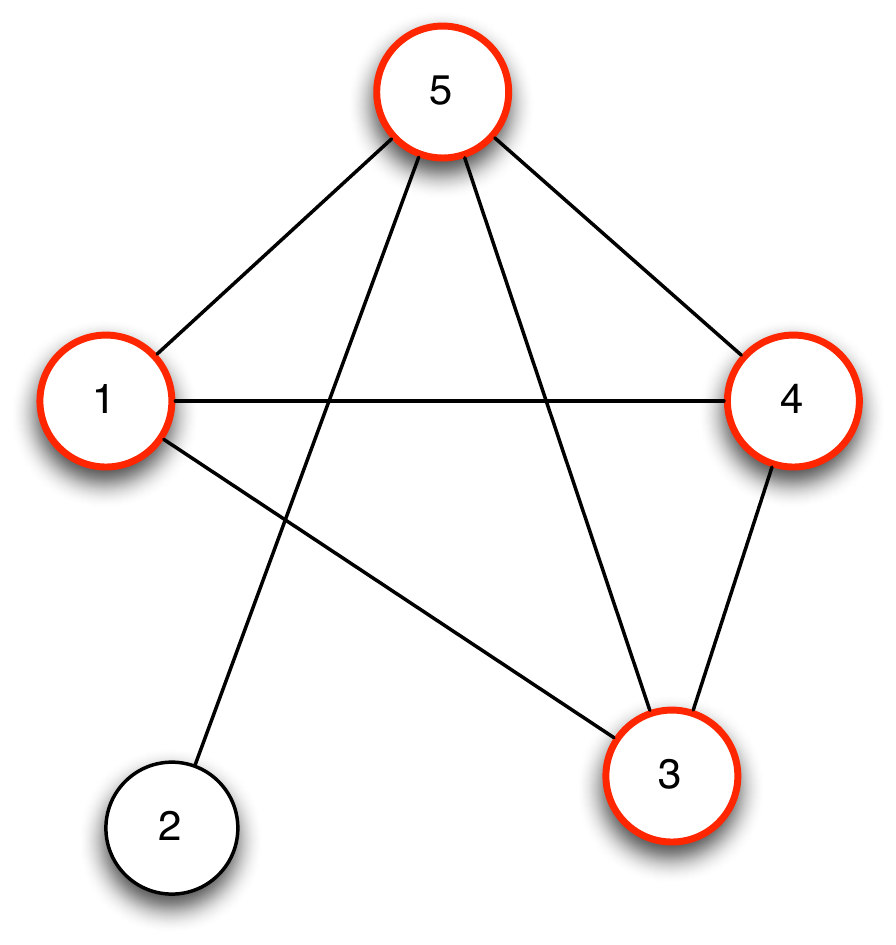}
	\caption{The maximum clique problem (MCP)}
	\label{fig:clique}
\end{figure}

Consider a set of kernel functions $\{\kappa(\cdot,\mr{\ell})\}_{\ell=1,\ldots,L}$. Determining a subset $\cp{D}$ with a prescribed coherence level can be viewed as a two-step procedure. The first step aims at listing all the pairs of functions that satisfy the coherence rule~\eqref{eq:coherenceThreshold}. This can be performed by constructing a $L \times L$ binary matrix $\adjM$ with entries defined as:
\begin{equation}
	\adjM_{ij} = \left\{
	\begin{array}{l l}
		1 & \,\, \text{if $\,\,|\kappa(\mr{i},\mr{j})| \leq \mu_0$}\\
		0 & \,\, \text{otherwise.}
	\end{array} \right. 
	\label{eq:Econd}
\end{equation}
The second step consists of finding in $\adjM$, up to a simultaneous reordering of its rows and columns, the largest submatrix of only ones. This problem can be recast as determining a maximum clique in an undirected graph $\cp{G}=\{V,E\}$, where each vertex $\ell$ of $V=\{1,\ldots,L\}$ corresponds to a candidate function $\kappa(\cdot,\mr{\ell})$, and edges in $E\subseteq V\times V$ connecting the vertices are defined by the adjacency matrix~$\adjM$. Two vertices are said to be adjacent if they are connected by an edge. A complete subgraph of $\cp{G}$ is one whose vertices are pairwise adjacent. The maximal clique problem (MCP) consists of finding the maximal complete subgraph of $\cp{G}$~\cite{pardalos1994maximum}. This problem is NP-Complete~\cite{karp1972reducibility}. Figure~\ref{fig:clique} illustrates this problem within the context of BS. This figure shows for instance that the coherence of $\kappa(\cdot,\mr{1})$ and $\kappa(\cdot,\mr{4})$ is lower than the preset threshold $\mu_0$, and the coherence of $\kappa(\cdot,\mr{1})$ and $\kappa(\cdot,\mr{2})$ is larger than $\mu_0$. This graph has one maximum clique defined by the set of vertices $\cp{I}_\cp{D}=\{1,3,4,5\}$, which means that the coherence of the dictionary $\cp{D}=\{\kappa(.,\mr{j})\}_{j\in\cp{I}_D}$ is lower than $\mu_0$ and it has maximum cardinality. A vast literature exists on maximum clique problems (MCP), see~\cite{wu2015review} and references therein. The next section reviews the main algorithms for MCP.

\subsection{The maximum clique problem}
 
MCP has a wide range of practical applications arising in a number of domains such as bioinformatics, coding theory, economics, social network analysis, etc. Given its theoretical importance and practical interests, considerable efforts have been devoted for deriving exact and heuristic algorithms. Efficient exact methods have been designed mainly based on the branch-and-bound (B\&B) framework. Dynamic bounds on the clique size are used to prune (or discard) branches during search, and then dramatically reducing the search space~\cite{carraghan1990exact}. Although algorithms are now much faster and efficient than their past counterparts~\cite{li2013combining}, the inherent complexity of exact methods can still lead to a prohibitive computation time when large problems are addressed~\cite{wu2015review}. To handle problems whose optimal solutions cannot be reached within a reasonable time, various heuristic and metaheuristic algorithms have been derived with the purpose of providing sub-optimal solutions in an acceptable time. In this paper, however, we shall focus on exact algorithms since our application concerns small graphs with a number of vertices equal to the number of bands.

% Many equivalent formulations as an integer programming problem, or as a continuous nonconvex optimization problem have been proposed for the MCP~\cite{pardalos1994maximum,butenko2003maximum}. Although different formulations shed light into different aspects of the problem, helping to prove new bounds for the clique size, and defining different strategies for the branching procedure, not all provided efficient approaches to solve the MCP~\cite{pardalos1994maximum}.
% The simplest formulation for the unweighted MCP is given by the following binary programming 
% \begin{equation}
%  \begin{split}
%   %C=&\mathop{\max\arg}_{\bx} \sum_{i =1 }^{L} x_i \\
%   &\max\sum_{i =1 }^{L} x_i \\
%   &\text{s.t. } x_i + x_j \leq 1,\quad \forall \{i,j\} \in \bar{E} \\
%   &x_i \in \{0,1\},\quad i=1,\ldots, L
%  \end{split}
% \end{equation}
% where the vertex $i$ is in the clique if $x_i=1$ and otherwise if $x_i=0$.

Since the introduction of the Carraghan and Pardalos (CP) exact algorithm~\cite{carraghan1990exact}, many refinements have been proposed to improve its performance with a focus on two main issues. The first one is to tighten the upper bound on the maximum clique during search for the purpose of more efficient subtree pruning. The second one is to improve the branching rule, and then select the most promising vertices to expand candidate cliques. In~\cite{wu2015review}, the authors classify the exact MCP algorithms into four groups, depending on their strategies for pruning and branching. The first group solves sub-clique problems for each vertex with iterative deepening and pruning strategies. Examples are the CP algorithm~\cite{carraghan1990exact} and its improved version~\cite{ostergaard2002fast}. Both algorithms are sensitive to the order of vertices, which can result in drastically different execution times for a given graph~\cite{ostergaard2002fast}. A second group is based on vertex coloring techniques~\cite{biggs1990some}. The most prominent algorithms in this group use B\&B strategies based on subgraph coloring. Examples of algorithms are BT and the recent MCQ, MCR, MaxCliqueDyn, BB-MaxClique, among others~\cite{wu2015review}. The third group improves the basic CP by tightening candidate sets via the removal of vertices that cannot be used to extend the current clique to a maximum clique. Along this line, three B\&B algorithms, denoted DF, $\chi$ and $\chi+$DF were proposed in~\cite{fahle2002simple}. The fourth group consists of the exact methods based on MaxSAT~\cite{li2010efficient}, which improve the techniques based on vertex coloring. The MaxCLQ algorithm proposed in~\cite{li2010efficient} is considered to be very effective and solved the DIMACS problem (p\_hat1000--3) for the first time~\cite{wu2015review}. A complex approach (ILS\&MaxCLQ) that combines different algorithms such as the MaxCLQ, MCS and the ILS, was recently proposed~\cite{maslov2014speeding}. A comparative discussion on exact methods is presented in~\cite{wu2015review}. The MaxCLQ and ILS\&MaxCLQ were the only methods to solve all the presented problems, with the smallest CPU times for the former.

\section{Algorithms}

We shall now introduce kernel BS algorithms based on the coherence criterion. As a baseline for performance comparisons, we consider first a greedy strategy that consists of testing candidate kernel functions sequentially and inserting them into the dictionary if coherence stays below a threshold value $\mu_0$. Next, we propose an exact strategy based on MCP solving.

\subsection{Automatic parameter settings}
\label{sec:ParSetting}

Before describing the kernel BS methods, we briefly present a procedure for automatic parameter setting. It allows to set the coherence threshold $\mu_0$ and Gaussian kernel bandwidth $\sigma^2$ given a desired number of elements in the dictionary.

Let $\bK\!_{\sigma}$ be the $L \times L$ Gram matrix whose $(i,j)$-th entry is defined by $\kappa_\sigma(\mr{i},\mr{j})$, where $\kappa_\sigma$ denotes the Gaussian kernel~\eqref{eq:gaussian-kernel} parameterized by the bandwidth $\sigma^2$. Let $\cp{D}$ be an $M$-element dictionary with coherence $\mu$ and index set $\cp{I}_{\cp{D}}$. Then, as shown in~\cite{Richard2009}, a sufficient condition for linear independence of the $M$ elements of $\cp{D}$ is given by $(M-1)\mu < 1$. We write:
\begin{equation}
	\mu< \frac{1}{(M-1)}.
\end{equation}
The objective is to build a dictionary with (approximately) $M$ linearly independent elements. We thus propose to set the coherence threshold $\mu_0$ as:
\begin{equation}
	\mu_0 = \frac{1}{(M-1)}
	\label{eq:mu_0}
\end{equation}
and adjust $\sigma^2$ to obtain a Gram matrix $\bK\!_{\sigma}$ whose entries are close to $\mu_0$ in some sense. Indeed, on the one hand, if all the off-diagonal entries of $\bK\!_{\sigma}$ are smaller than $\mu_0$, then $\cp{D}$ contains the $L$ available elements. On the other hand, if all the off-diagonal entries of $\bK\!_{\sigma}$ are greater then $\mu_0$, then $\cp{D}$ should be composed of only one element. Therefore, we propose to adjust $\sigma^2$ such that $\mathbb{E}\{(\bK\!_{\sigma_{ij}})_{(i\neq j)}\} = \mu_0$, where $\mathbb{E}\{\cdot\}$ is the expected value and can be approximated as 

% 
% \bigskip
% \bigskip
\begin{equation}
\label{eq:KExpValue}
\mathbb{E}\{(\bK\!_{\sigma_{ij}})|_{(i\neq j)}\} \approx \frac{2}{L^2-L}\sum_{i=1}^{L-1}\sum_{j=i+1}^{L}\bK\!_{\sigma_{ij}}. 
\end{equation}

Then, we set $\sigma^2$ as the solution of the following optimization problem: 
\begin{equation}
\begin{split}
 \sigma^2 =& \mathop{\arg\min}_{\sigma^2} \left( \frac{2}{L^2-L}\sum_{i=1}^{L-1}\sum_{j=i+1}^{L} [\bK_{1_{ij}}]^{1/\sigma^2} - \mu_0\right)^2\\
 & \text{s. t.}\quad \sigma^2\in\mathbb{R}^+.
\end{split}
\label{eq:sigmaOpt}
\end{equation}
where $\bK_{1} = \bK\!_{\sigma}$ is the Gram matrix for $\sigma = 1$. Finally, we determine $\bK\!_{\cp{D}}$ as the largest sub-matrix of $\bK\!_{\sigma}$ whose all off-diagonal entries satisfy~\eqref{eq:coherenceThreshold}. We emphasize that since $\bK\!_{\sigma_{ij}}\leq 1$, \eqref{eq:KExpValue} is a decreasing function of $\sigma^{-2}$, and thus~\eqref{eq:sigmaOpt} has a unique solution.

\subsection{Algorithms}
In this section we present the two band selection algorithms using the greedy and clique approaches that will be used in Section~\ref{sec:Application}.

The greedy coherence-based approach is presented in Algorithm~\ref{alg:GCBS}. The inputs to Algorithm~\ref{alg:GCBS} are the desired number $M$ of bands in the final dictionary, and the $L\times L$ Gaussian kernel Gram matrix with $\sigma = 1$ and entries $\bK_{1_{ij}}=\kappa(\mr{i},\mr{j})=\exp\left(-0.5\|\mr{i}-\mr{j}\|^2\right)$. It returns the index of selected bands and the the Gaussian kernel bandwidth $\sigma^2$. Initialization occurs in line 1, where the index set $\cp{I_D}$ is initialized with the first spectral band index, the number $N_b$ of bands in the dictionary is set to one, and the coherence threshold $\mu_0$ is adjusted according to~\eqref{eq:mu_0}. Next, $\sigma^2$ is determined by solving problem~\eqref{eq:sigmaOpt} in line 2, and the Gram matrix $\bK_{\sigma}$ is computed with the optimum $\sigma^2$ in line 3. From line 4 to line 13 the algorithm sequentially tests all the $L-1$ remaining bands using condition~\eqref{eq:coherenceThreshold}. Breaking the parts down, in line 5 a zero vector $\bc$ of length $N_b$ is created, and the off diagonal terms $(\ell, \cp{I}_{\cp{D}_j})$ of the Gram matrix $\bK_{\sigma}$ are stored in $\bc$. If the maximum absolute value of the entries of $\bc$ is less than the coherence threshold (line 9), then the $\ell$-th band index is added to $\cp{I}_{\cp{D}}$, and $N_b$ is incremented by one (lines 10 and 11). Finally, the algorithm returns the complete set of selected bands and the kernel bandwidth in line 14.

\begin{algorithm}
\SetKwInOut{Input}{Input}
\SetKwInOut{Output}{Output}
\caption{Greedy Coherence-based Band Selection (GCBS)~\label{alg:GCBS}}
\Input{The $L\times L$ Gram matrix $\bK_1= (\bK_{\sigma})_{\sigma=1}$, and the desired number $M$ of atoms.}
\Output{The indices $\cp{I_D}$ of selected atoms, and the Gaussian kernel bandwidth $\sigma^2$.}
%Initialization: $\cp{D} = \{1\}$, $N_b= 1$\; 
Initialization: $\cp{I_D} = \{1\}$, $N_b= 1$, $\mu_0 = 1/(M-1)$\; 
Find $\sigma^2$ solving~\eqref{eq:sigmaOpt}\;
Compute $\bK_{\sigma}$ using $\sigma^2$ obtained in line 2\;
\For{$\ell:=2$ \KwTo $L$}{  
    $\cb{c}:= \cb{0}_{N_b\times 1}$\; 
    \For{$j:=1$ \KwTo $N_b$}{ 
	$\cb{c}_j := \bK_{\sigma_{\ell,\cp{I_D}_j}}$ \;
     }
      \If {$\max (|\cb{c}_j|) \leq \mu_0$ }{
	Insert $\ell$ into $\cp{I_D}$\;
	$N_b := N_b + 1$\;
      }  
 }
 
 \KwRet $\cp{I_D}$, $\sigma^2$\;
\end{algorithm}

The clique coherence-based band selection method is described in Algorithm~\ref{alg:CCBS}. Similarly to Algorithm~\ref{alg:GCBS}, the inputs are $\bK_{1}$ and $M$.  The adjacency matrix $\cb{B}$ in initialized with zeros (line 1), the vertices vector $V$ with the indices of all available wavelengths, $\mu_0$ following \eqref{eq:mu_0}, and $\cp{I}_{\cp{D}}$ as an empty set. The kernel bandwidth is computed in line 2, and the Gram matrix is computed for the optimum $\sigma^2$ in line 3. Through line 4 to 10 every entry of the upper diagonal part of $\cb{B}$ is set according to~\eqref{eq:Econd}. In line 11 the \emph{MaxCLQ} algorithm is used to find the indices of the maximum clique in the graph. These indices are assigned to the dictionary index set $\cp{I_D}$, which is returned in line 10 together with the kernel bandwidth.

\begin{algorithm}
\SetKwInOut{Input}{Input}
\SetKwInOut{Output}{Output}
\caption{Clique Coherence-based BS (CCBS)~\label{alg:CCBS}}
\Input{The $L\times L$ Gram matrix $\bK_1= (\bK_{\sigma})_{\sigma=1}$, and the desired number $M$ of atoms.}
\Output{The indices $\cp{I_D}$ of selected atoms, and the Gaussian kernel bandwidth $\sigma^2$.}
%Initialization: $\cp{D} = \{1\}$, $N_b= 1$\; 
Initialization: $\cb{B}:= \cb{0}_{L\times L}$,  $V=\{1,\ldots, L\}$, $\mu_0 = 1/(M-1)$, $\cp{I_D}_c=\{\emptyset\}$\; 
Find $\sigma^2$ solving~\eqref{eq:sigmaOpt}\;
$\bK_{\sigma}$ using $\sigma^2$ obtained in line 2\;
\For{$i:=1$ \KwTo $L-1$}{      
    \For{$j:=i+1$ \KwTo $L$}{ 
	%\If {$[\bK_{i,j}]^{1/\sigma^2}\leq \mu_0$ }{
	\If {$[\bK_{\sigma_{ij}}]\leq \mu_0$ }{
	  $\cb{B}_{ij} := 1$\;
	}
	
     }
  }
 $\cp{I_D} := \textit{MaxCLQ}(V,\cb{B})$\;
 
 \KwRet $\cp{I_D}$, $\sigma^2$\;
\end{algorithm}

Note that $M$ is used in Algorithm~\ref{alg:CCBS} and Algorithm~\ref{alg:GCBS} as a design parameter, which is required to obtain the coherence threshold and the Gaussian kernel bandwidth. The number $N_b$ of bands in the final dictionary can differ from $M$.

\section{Application}\label{sec:Application}

\subsection{The SK-Hype}

This section reviews the SK-Hype algorithm\footnote{Matlab code available at www.cedric-richard.fr} for nonlinear unmixing of HIs~\cite{Chen-2013-ID321}. It considers the mixing model consisting of a linear trend parametrized by the abundance vector $\balpha$ and a nonlinear residual component $\psi$.  This model is given by
\begin{equation}
\label{eq:nlmodel}
%r_\ell = u\,\balpha^\top \bm_{\lambda_\ell} + (1-u)\,\psi(\bm_{\lambda_\ell}) + n_\ell
r_\ell = u\,\balpha^\top \mr{\ell} + (1-u)\,\psi(\mr{\ell}) + n_\ell
\end{equation}
where $u\in [0,1]$ controls the amount of linear contribution to the model and $\psi(\cdot)$ is an unknown function in an RKHS $\cp H$. 
%
% Kernel methods are efficient machine learning techniques that were initially introduced for solving nonlinear classification and regression problems. They consist of linear algorithms operating in high dimensional feature spaces into which the data have been mapped using kernel functions~\cite{Vapnik1995}. These approaches are based on the framework of reproducing kernels which states that, for any positive kernel $\kappa(\mr{i},\mr{j})$, there exists a Hilbert space $\cp H$ with inner product $\psh{\cdot\,}{\cdot}$ and a mapping
% \begin{eqnarray}
% 	{\bf\Phi}:\quad {\mathbb R}^L&\longrightarrow & H \\
% 	 \mr{i} &\longmapsto& \kappa(\cdot,\mr{i})
% \end{eqnarray}
% such that $\kappa(\mr{i},\mr{j}) = \psh{\bf\Phi(\mr{i})}{\bf\Phi(\mr{j})}$. This last property allows to implicitly compute inner products in $\cp H$ by evaluating a real function, $\kappa(\mr{i},\mr{j})$, in the input space. Other useful properties are $\psi(\bm_{\lambda_j})=\langle\psi,\kappa(\cdot,\bm_{\lambda_j})\rangle_\cp{H}$ for all $\psi$ in $\cp{H}$, and the reproducing property 
% $\kappa(\bm_{\lambda_i},\bm_{\lambda_j})=\langle\kappa(\cdot,\bm_{\lambda_i}),\kappa(\cdot,\bm_{\lambda_j})\rangle_\cp{H}$.
%
SK-Hype solves the optimization problem
%proposed in~\cite{Chen-2013-ID321} for estimating the unknown variables $\balpha$, $\psi(\cdot)$ and $u$ in \eqref{eq:nlmodel} is
%  \begin{equation}
%  \label{eq:optproblem}
%  \begin{split}
% \min_{\balpha,\psi,u} \frac{1}{2}&\left(\frac{1}{u}\|\balpha\|^2+\frac{1}{1-u}\|\psi\|_\cp{H}^2\right)\\
% &+\frac{1}{2\mu}\sum_{\ell=1}^L (r_\ell-\balpha^\top\mr{\ell}-\psi(\mr{\ell}))^2
% \end{split}
%  \end{equation}
\begin{equation}
 \label{eq:optproblem}
 \begin{split}
&\min_{\balpha,\psi,u} \frac{1}{2}\left(\frac{1}{u}\|\balpha\|^2+\frac{1}{1-u}\|\psi\|_\cp{H}^2\right) +\frac{1}{2\mu}\sum_{\ell=1}^L e_\ell^2\\
&\text{subject to}\quad \balpha \succeq \bf{0}, \, \cb{1}^\top\balpha = 1,\text{ and } \\
&\qquad\qquad\quad e_\ell = r_\ell - u\,\balpha^\top \mr{\ell} - (1-u)\,\psi(\mr{\ell}).
\end{split}
 \end{equation}
which is convex under mild continuity conditions~\cite{Chen-2013-ID321}. Problem \eqref{eq:optproblem} is solved using a two stage alternating iterative procedure with respect to $(\balpha,\psi)$ and $u$. For fixed $u$ and Lagrange multipliers $\bbeta$ and $\bgamma$, the dual problem of~\eqref{eq:optproblem} is given by~\cite{Chen-2013-ID321} 
\begin{equation}
	\label{eq:dual.algo2}
	\begin{split}
      		\max_{\bbeta,\bgamma}\,\, &G(u,\bbeta,\bgamma)  =\\ 
		&-\frac{1}{2}\left(
			\begin{array}{c}
				\bbeta \\ \hline
 				\bgamma
			\end{array}\right)^{\!\!\!\top}
			\left(
			\begin{array}{c|c}
				\bK_{u}+\mu\bI  & u\MM \\ \hline
 				u\MM^\top & u\bI
			\end{array}\right)
			\left(
			\begin{array}{c}
				\bbeta \\ \hline
 				\bgamma
			\end{array}\right)\\
			&+ \left(
			\begin{array}{c}
				\br \\ \hline
 				\bf{0}
			\end{array}\right)^{\!\!\!\top}
			\left(
			\begin{array}{c}
				\bbeta \\ \hline
 				\bgamma
			\end{array}\right)
			\\
      		& \text{subject to} \quad \bgamma \succeq \bf{0}
	\end{split}
\end{equation}
with $\bK_{u} = u\MM\MM^\top + (1-u)\KK$. Solving~\eqref{eq:dual.algo2} is equivalent to solving the linear system 
\begin{equation}
  \label{eq:dualSystemSKHYPE}
  \left(
     \begin{array}{c|c|c}
       -\bI & \bK_u + \mu\bI & u\MM\\\hline
       \cb{0} & u\MM^\top & u\bI
     \end{array}\right)
  \left(
     \begin{array}{c}
       \br \\ \hline
       \bbeta\\\hline
       \bgamma
     \end{array}\right)			
  =
  \cb{0}.
\end{equation}
Denoting $\bbeta^*$ and $\bgamma^*$ the solutions of \eqref{eq:dual.algo2}, the solution of the primal problem \eqref{eq:optproblem} for $u$ fixed is~\cite{Chen-2013-ID321}
\begin{equation}
	\label{eq:solution.algo2}
	\left\{
    		\begin{array}{ll}
    			\balpha^* = \frac{\MM^\top\bbeta^*+\bgamma^*}{\cb{1}^\top(\MM^\top\bbeta^*+\bgamma^*)} \\
    			\psi^* = (1-u)\sum_{\ell=1}^L \beta_\ell^*\,\kappa(\cdot,\mr{\ell}) \\					 
        			e_\ell^* = \mu\,\beta_\ell^*
    	\end{array}
	\right.
\end{equation}
The alternating optimization is completed by using \eqref{eq:solution.algo2} in~\cite{Chen-2013-ID321}, defining the resulting cost function $J(u)$, solving 
\begin{equation}
\label{eq:solutionJu}
 \min_{u} J(u) \quad\text{subject to}\quad 0<u<1
\end{equation}
and continue by iteratively solving \eqref{eq:dualSystemSKHYPE} and \eqref{eq:solutionJu} to find the global solution~\cite{Chen-2013-ID321}.

%Once $\bbeta^\star$ and $\bgamma^\star$ are determined, pixel reconstruction can be performed using $\br^\star = [\psi^\star(\mr{1}), \ldots, \psi^\star(\mr{L})]^\top$ with $\psi^\star(\mr{\ell})=\bm_{\lambda_\ell}^\top\,\bh^\star+\psi_\text{nlin}^\star(\bm_{\lambda_\ell})$ defined in~\eqref{eq:solution.algo2}. 
%Finally, the estimated abundance vector is given by 
%\begin{equation}
%	\label{eq:alpha.algo2}
%	\balpha^\star = \frac{\MM^\top\bbeta^\star+\bgamma^\star}{\cb{1}^\top(\MM^\top\bbeta^\star+\bgamma^\star)}.
%\end{equation}

% \subsubsection{Solving with respect to $u$}
% Using the stationary conditions~\eqref{eq:solution.algo2}, the optimum value for $u$ given $\bbeta^\star$ and $\bgamma^\star$ can be computed each iteration as~\cite{Chen-2013-ID317}
% \begin{equation}
% 	{ u^\star = \left( 1+ (1-u^\star_{-1})\sqrt{\frac{{\bbeta^\star}^\top\bK\bbeta^\star}{\|\MM^\top\bbeta^\star+\bgamma^\star \|^2}}\right)}
% \end{equation}
% where $u^\star_{-1}$ is the optimum $u^\star$ obtained at the previous iteration.

\subsection{Simulation with synthetic data}\label{sec:SimSynthetic}
\begin{figure}[ht]
 \centering
 \includegraphics[width=0.5\textwidth]{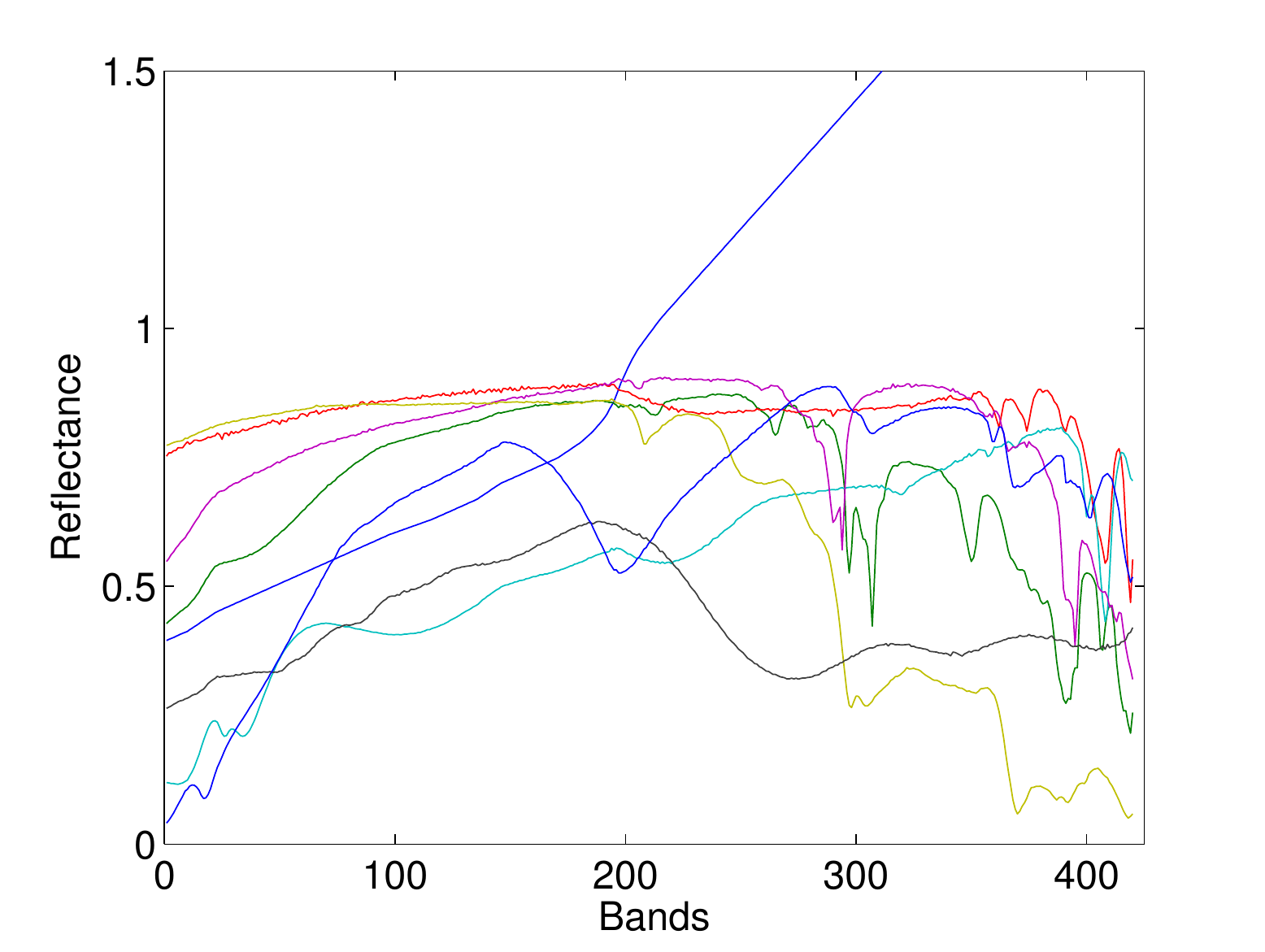}
 \caption{Eight Cuprite minerals reflectances extracted from the {ENVI} software.}\label{fig:cupEM}
\end{figure}

\begin{figure}[ht]
 \centering
 \includegraphics[width=0.5\textwidth]{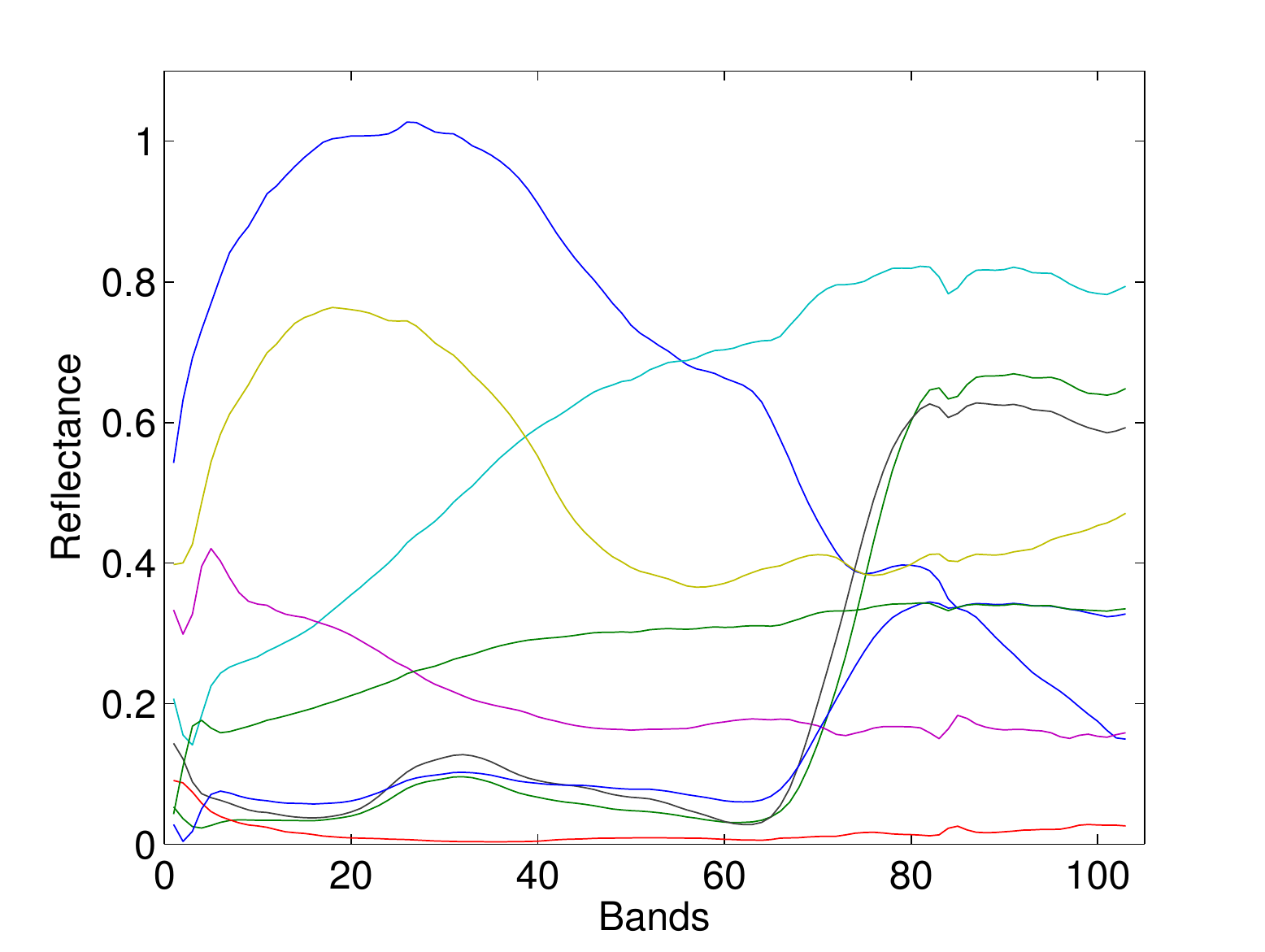}
 \caption{Nine Pavia University endmembers extracted using VCA.}\label{fig:pavEM}
\end{figure}

\begin{figure}
\captionsetup[subfigure]{position=b}
         \centering
        %\begin{subfigure}[b]{0.3205\textwidth}
        \begin{subfigure}[b]{0.21\textwidth}
                \includegraphics[width=\textwidth]{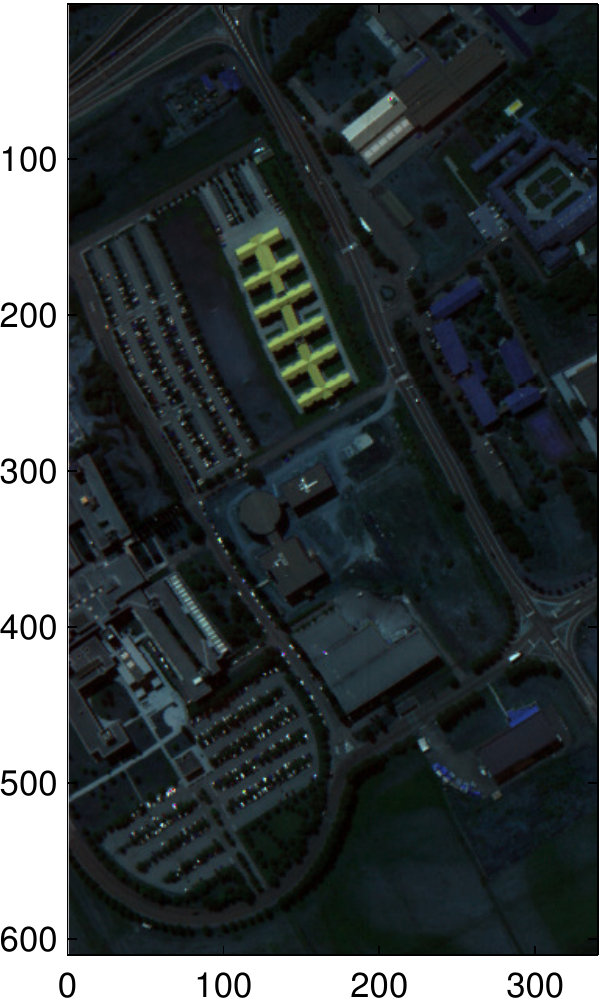}
                \caption{Pavia University representation.}
                \label{fig:paviaU}
        \end{subfigure} %\vspace{0.5cm}
         %\qquad %add desired spacing between images, e. g. ~, \quad, \qquad, \hfill etc.
        %\begin{subfigure}[b]{0.4\textwidth}
        \begin{subfigure}[b]{0.263\textwidth}
                \includegraphics[width=\textwidth]{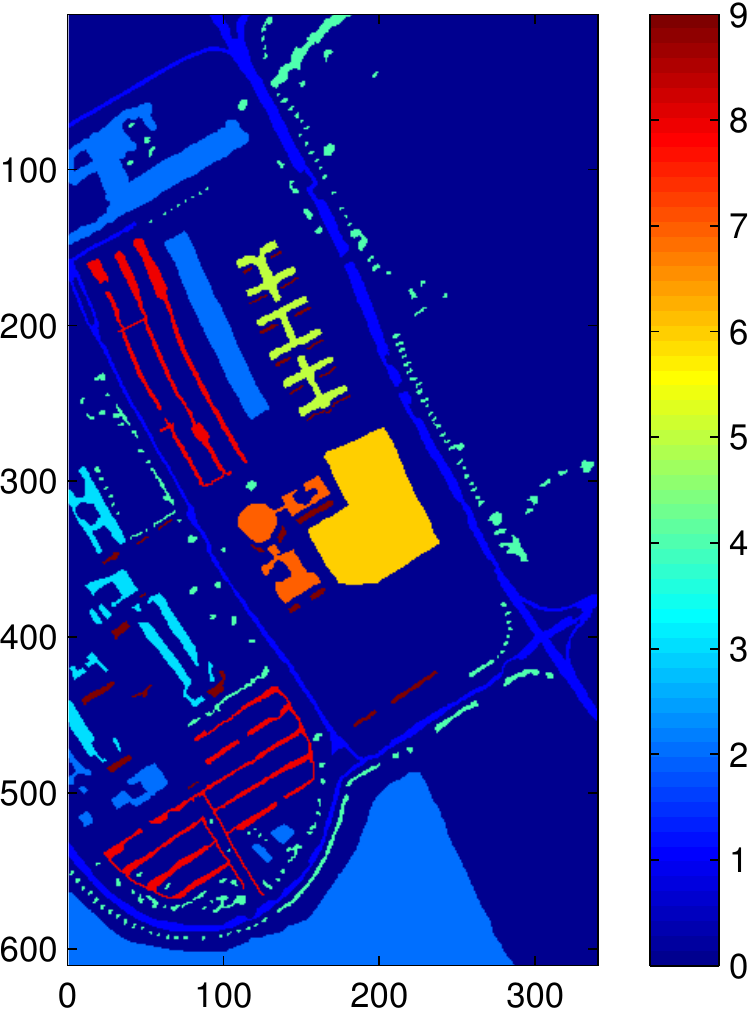}
                \caption{Ground truth for the Pavia University scene.}
                \label{fig:paviaU_gt}
        \end{subfigure}
        \caption{Pavia University. In (a) the Pavia University HI is represented using the bands 5, 30, and 50. In (b) the classified areas are labelled from 1 to 9, while 0 corresponds to unclassified areas.}\label{fig:pavia}
\end{figure}

This section presents simulation results using synthetic data to illustrate the performance of the proposed  unmixing method under controlled conditions for which the abundance values are known. We constructed synthetic images using two sets of endmembers. The first set had 8 endmembers extracted from the spectral library of the {ENVI} software and correspond to the spectral signatures of minerals present in the Cuprite mining field in Nevada. The minerals are alunite, calcite, epidote, kaolinite, buddingtonite, almandine, jarosite and lepidolite, and their spectra consisted of 420 contiguous bands, covering wavelengths from $0.3951\mu$m to $2.56\mu$m, and their reflectances are displayed in Figure~\ref{fig:cupEM}. 
The second set was extracted from the Pavia University data acquired by the ROSIS spectrometer. It has $610\times340$ pixels with 103 bands over the spectral range of 430--680 nm (Figure~\ref{fig:paviaU}). The data also has a ground truth labelling 42776 pixels (out of the 207400) into 9 classes labeled asphalt, meadows, gravel, trees, painted metal sheets, bare soil, bitumen, self-blocking bricks and shadows (Figure~\ref{fig:paviaU_gt}). We extracted the endmembers from this data set using the vertex component analysis algorithm (VCA~\cite{Nascimento2005}), and considering only the labeled pixels. The reflectances for the 9 endmembers extracted with VCA are showed in Figure~\ref{fig:pavEM}.
We constructed four 2000-pixel hyperspectral images ($N=2000$), each using 8 endmembers ($R=8$) from the Cuprite or Pavia data, and the simplified GBM or PNMM mixing models (see Section~\ref{sec:HI}) with $\delta = 1$ and $\xi=0.7$, respectively. The abundances were obtained by uniformly sampling from the simplex, i.e., obeying the positivity and sum-to-one constraints. WGN was added to all images with power adjusted to produce a 21dB SNR. We consider the root mean square error (RMSE) in abundance estimation
\begin{equation}
	\text{RMSE}=\sqrt{\frac{1}{NR} \sum_{n=1}^{N}\|\balpha_n-\balpha^*_n\|^2}
	%\|\bA-\widehat{\!\bA}\|_{F}/(NR)
\end{equation}
and the CPU time required for both BS (when applicable) and unmixing (averaged over 100 unmixings of the same HIs) to compare the different BS strategies.  All unmixings were performed using a Gaussian kernel and considering either the full set of bands or smaller sets selected using the BS strategies presented in Section~\ref{sec:BS}. SK-Hype was implemented for the full set of bands. The kernel bandwidth for SK-Hype was selected among the values $\sigma_{\text{skp}} \in \{0.5\sigma,\, \sigma,\, 2\sigma,\, 10\sigma,\,20\sigma\}$ to obtain the minimum RMSE, where $\sigma$ is the solution of~\eqref{eq:sigmaOpt}, for $M=30$. The global kernel k-means (GKKM) algorithm~\cite{Imbiriba2015} implementation requires the number of bands to be fixed \emph{a priori}.  We considered a selection approach based on the Akaike Information Criterion and given by~\cite{manning2008introduction}
\begin{equation}
 M = \mathop{\arg\min}_{M}[ E(\nu_1,\ldots,\nu_M) + \lambda M]
\end{equation}
where the parameter $\lambda$ controls the complexity of the model, and needs to be found empirically. The kernel bandwidth $\sigma_\text{kkm}$ also needs to be selected for GKKM. A grid search was performed using a small part (200 pixels) of the synthetic image to find $\lambda$ and $\sigma_\text{kkm}$ that would lead to a good RMSE performance. The parameters were chosen among the values $\lambda \in \{2,  4, 6\}$ and $\sigma_{\text{kkm}} \in \{0.5\sigma,\, \sigma,\, 2\sigma,\, 10\sigma,\,20\sigma\}$, again with $\sigma$ being the solution of~\eqref{eq:sigmaOpt}, for $M=30$. The parameter set leading to the best performance in terms of RMSE for the abundances was then selected. It is important to notice that, in general, the abundance ground truth is not available from real data. Thus, the RMSE in abundance estimation could not be used in design as a measure to select model parameters. Hence, the SK-Hype and GKKM designs used in this comparison are based on a quasi-optimal choice of parameters for these methods, which could not be determined in practice. The proposed design for the BS methods, however, can be employed in practical applications.
%~\footnote{\cred{Note that the abundance ground truth are not available in real data and here are used with the only purpose of assessing the potential of SK-Hype and GKKM}.}.

BS with the CCBS and GCBS algorithms was performed using $M \in \{5,\,10,\,20,\,30\}$, with parameters $\mu_0$ and $\sigma$ adjusted using the methodology presented in Section~\ref{sec:ParSetting}. 
We emphasize that this parameter setting strategy assumes no \emph{prior} knowledge about the abundance ground truth.

%As the global kernel k-means (GKKM) algorithm~\cite{Imbiriba2015} requires the number of bands to fixed \emph{a priori}, we have set $N_b$ for the GKKM equal to the dictionary size obtained using the proposed CCBS method, and not $M$, to facilitate the comparison\footnote{Note that $N_b$ must be guessed for GKKM, while it is optimized in CCBS.}. Moreover, we have used the same kernel parameter selected by the CCBS algorithm for each $M$ in implementing GKKM. Hence, the presented comparison between GKKM and the BS methods is based solely on the achievable dictionary coherence. 
%For the pure SK-Hype implementation using all bands, the kernel bandwidth was chosen as the one leading to the best performance among 5 possible values $\sigma_{\text{skp}} \in \{0.5\sigma_\star,\, \sigma_\star,\, 2\sigma_\star,\, 10\sigma_\star,\,20\sigma_\star\}$, where $\sigma_\star$ is the kernel bandwidth obtained using~\eqref{eq:sigmaOpt} and $M=30$. 

The simulation results are summarized in Tables~\ref{tab:CupriteSynthetic_Diff_m_gbm} to  \ref{tab:PaviaSynthetic_PNMM}. In these tables, the first column shows the BS strategy considered prior to unmixing. SK-Hype in this column indicates the solution without BS. The symbol "(r)" besides CCBS or GCBS means that we have randomized the order of the bands prior to applying the BS strategy. The second column shows the obtained RMSE and the standard deviation (STD) in abundance estimation. The third column lists the average CPU time elapsed in the (BS + unmixing) process. Column four shows the number of selected bands $N_b$, and last column shows the coherence of the final dictionary. 

Tables~\ref{tab:CupriteSynthetic_Diff_m_gbm} and~\ref{tab:CupriteSynthetic_Diff_m_pnmm} show the results for HIs built with Cuprite endmembers and using, respectively, the GBM and the PNMM mixing models. Note that the RMSE obtained using the BS algorithms are very close to those obtained using all bands. Nevertheless the reduction in number of bands obtained through BS is at least tenfold.  The computational complexity advantage of the BS methods is evidenced by the required average CPU time, which show reductions by factors ranging from 50 to 110, depending on the algorithm and parameter settings. Note also that the number of bands in the final dictionary tends to be larger than the value $M$ used to initialize the algorithms. This increase in the anticipated number of bands is obtained to optimize the dictionary coherence, what is not possible in the GKKM algorithm. As expected, the number of bands remained the same for the clique algorithm (CCBS) for each value of $M$, and the slight changes in the RMSE results indicate that the maximum clique is not unique. For the greedy approach (GCBS), however, different numbers of bands are obtained at each execution due to initial randomization, and the results in terms of RMSE and CPU time vary slightly. In general, randomization did not have any significant impact on the results. Finally, one should note from these tables that the coherence-based algorithms produced dictionaries with coherence close to $\mu_0$, and 2 to 23 times smaller than the coherence obtained using GKKM.

%Results similar to those described above were obtained using the HIs created with data extracted from the Pavia database.  These results are described in detail in the extended report \cred{[Technical Report]}.

Tables~\ref{tab:PaviaSynthetic_GBM} and~\ref{tab:PaviaSynthetic_PNMM} show the results for the HIs created with the Pavia endmembers using the GBM and PNMM respectively. Although the results in Tables~\ref{tab:PaviaSynthetic_GBM} and~\ref{tab:PaviaSynthetic_PNMM} follows the same pattern that the results in Tables~\ref{tab:CupriteSynthetic_Diff_m_gbm} and~\ref{tab:CupriteSynthetic_Diff_m_pnmm}, we highlight that for the Pavia HIs the number of available bands is 103 in contrast to the 420 used in the previous example. This explains the smaller improvement in the Av. Time when using the BS algorithms which is about 3 to 4 times smaller than using all the bands. Another difference in the results is that using the BS algorithms, and its reasoning for setting $\mu_0$ and $\sigma^2$, the best results in terms of RMSE were obtained by the proposed method CCBS with $M=30$ in both Tables. When concerning the number of bands, the final $N_b$ were closer to $m$ than in the previous example. For the coherence of the final dictionary the same pattern obtained in Tables~\ref{tab:CupriteSynthetic_Diff_m_gbm} and~\ref{tab:CupriteSynthetic_Diff_m_pnmm} repeats for the Pavia HIs.

%
%  Code used for the simulations in the tables below:
%  Dropbox/matlab/HyperspectralCode/Test/TestCoherenceBSInitializations_Test_Heuristic.m

\begin{table}
\caption{RMSE. 100 runs, 2000 pxl., 8 endmembers (Cuprite), SNR=21dB, GBM, SK-Hype. $\mu_0$ computed using Equation~\eqref{eq:mu_0} for a given $M$, and $\sigma$ is found solving problem~\eqref{eq:sigmaOpt}.}
\label{tab:CupriteSynthetic_Diff_m_gbm}
\begin{center}
\begin{tabular}{|l|c|c|c|c|}\hline
Strategy & RMSE $\pm$ STD &  Av. Time & $N_b$ & $\mu$\\ \hline
%SK-Hype ($\sigma = 0.0177$) & 0.0857 $\pm$ 0.0037 &  17.26 $\pm$ 0.67 & 103\\
SK-Hype & 0.0680 $\pm$ 0.0028 &  301.08 $\pm$ 17.93 & 420 & -\\
\hline
%GKKM & 0.0667 $\pm$ 0.0026 &  17.14 $\pm$ 0.04 & 29 & 0.4004\\
GKKM & 0.0664 $\pm$ 0.0026 &  25.40 $\pm$ 0.22 & 36 & 0.5893\\
\hline
\multicolumn{5}{|c|}{$M=5$, $\mu_0 = 0.25$, $\sigma = 0.2548$}  \\ \hline
CCBS & 0.0687 $\pm$ 0.0028 &  3.10 $\pm$ 0.14 & 10 & 0.2482\\
CCBS (r) & 0.0687 $\pm$ 0.0028 &  3.13 $\pm$ 0.12 &  10 & 0.2482\\
GCBS & 0.0724 $\pm$ 0.0031 &  2.91 $\pm$ 0.02 & 8 & 0.2482\\
GCBS (r) & 0.0721 $\pm$ 0.0030 &  3.15 $\pm$ 0.15 & 7.13 $\pm$ 0.97 & 0.2331\\
\hline
\multicolumn{5}{|c|}{$M=10$, $\mu_0 = 0.1111$, $\sigma = 0.1320$}  \\ \hline
CCBS & 0.0678 $\pm$ 0.0027 &  2.85 $\pm$ 0.13 & 16 & 0.1108\\
CCBS (r) & 0.0679 $\pm$ 0.0027 &  2.89 $\pm$ 0.17 &  16 & 0.1108\\
GCBS & 0.0685 $\pm$ 0.0028 &  2.57 $\pm$ 0.02 & 16 & 0.1104\\
GCBS (r) & 0.0688 $\pm$ 0.0028 &  2.65 $\pm$ 0.06 & 13.09 $\pm$ 1.10 & 0.0996\\
\hline 
\multicolumn{5}{|c|}{$M=20$, $\mu_0 = 0.0526$, $\sigma = 0.0965$}  \\ \hline
CCBS & 0.0659 $\pm$ 0.0026 &  2.96 $\pm$ 0.15 & 21 & 0.0520\\
CCBS (r) & 0.0660 $\pm$ 0.0026 &  3.01 $\pm$ 0.17 &  21 & 0.0520\\
GCBS & 0.0670 $\pm$ 0.0027 &  2.59 $\pm$ 0.02 & 20 & 0.0525\\
GCBS (r) & 0.0678 $\pm$ 0.0027 &  2.67 $\pm$ 0.08 & 15.95 $\pm$ 1.13 & 0.0467\\
\hline 
\multicolumn{5}{|c|}{$M=30$, $\mu_0 = 0.0345$, $\sigma = 0.0503$}  \\ \hline
CCBS & 0.0637 $\pm$ 0.0024 &  5.54 $\pm$ 0.22 & 42 & 0.0339\\
CCBS (r) & 0.0637 $\pm$ 0.0024 &  5.74 $\pm$ 0.18 &  42 & 0.0339\\
GCBS & 0.0637 $\pm$ 0.0024 &  3.32 $\pm$ 0.04 & 41 & 0.0344\\
GCBS (r) & 0.0644 $\pm$ 0.0025 &  2.83 $\pm$ 0.07 & 33.39 $\pm$ 1.43 & 0.0326\\
\hline
\end{tabular}
\end{center}
\end{table}

\begin{table}
\caption{RMSE. 100 runs, 2000 pxl., 8 endmembers (Cuprite), SNR=21dB, PNMM, SK-Hype. $\mu_0$ computed using Equation~\eqref{eq:mu_0} for a given $M$, and $\sigma$ is found solving problem~\eqref{eq:sigmaOpt}.}
\label{tab:CupriteSynthetic_Diff_m_pnmm}
\begin{center}
\begin{tabular}{|l|c|c|c|c|}\hline
Strategy & RMSE $\pm$ STD &  Av. Time & $N_b$ & $\mu$\\ \hline
SK-Hype & 0.0728 $\pm$ 0.0030 &  277.03 $\pm$ 4.30 & 420 & -\\
\hline
GKKM & 0.0729 $\pm$ 0.0030 &  25.52 $\pm$ 0.18 & 36 & 0.7760\\
%GKKM & 0.0746 $\pm$ 0.0030 &  23.17 $\pm$ 0.53 & 29 & 0.4004\\
%GKKM & 0.0749 $\pm$ 0.0031 &  10.22 $\pm$ 0.07 & 21 & 0.6235\\
\hline
\multicolumn{5}{|c|}{$M=5$, $\mu_0 = 0.25$, $\sigma = 0.2548$}  \\ \hline
CCBS & 0.0748 $\pm$ 0.0031 &  2.99 $\pm$ 0.10 & 10 & 0.2482\\
CCBS (r) & 0.0749 $\pm$ 0.0031 &  3.12 $\pm$ 0.18 &  10 & 0.2482\\
GCBS & 0.0764 $\pm$ 0.0032 &  2.85 $\pm$ 0.06 & 8 & 0.2482\\
GCBS (r) & 0.0776 $\pm$ 0.0033 &  2.99 $\pm$ 0.15 & 7.13 $\pm$ 0.97 & 0.2331\\
\hline
\multicolumn{5}{|c|}{$M=10$, $\mu_0 = 0.1111$, $\sigma = 0.1320$}  \\ \hline
CCBS & 0.0746 $\pm$ 0.0031 &  2.85 $\pm$ 0.19 & 16 & 0.1108\\
CCBS (r) & 0.0745 $\pm$ 0.0031 &  2.84 $\pm$ 0.14 &  16 & 0.1108\\
GCBS & 0.0757 $\pm$ 0.0032 &  2.57 $\pm$ 0.04 & 16 & 0.1104\\
GCBS (r) & 0.0757 $\pm$ 0.0031 &  2.64 $\pm$ 0.10 & 13.09 $\pm$ 1.10 & 0.0996\\
\hline 
\multicolumn{5}{|c|}{$M=20$, $\mu_0 = 0.0526$, $\sigma = 0.0965$}  \\ \hline
CCBS & 0.0735 $\pm$ 0.0029 &  2.87 $\pm$ 0.12 & 21 & 0.0520\\
CCBS (r) & 0.0737 $\pm$ 0.0029 &  2.96 $\pm$ 0.17 &  21 & 0.0520\\
GCBS & 0.0753 $\pm$ 0.0031 &  2.55 $\pm$ 0.03 & 20 & 0.0525\\
GCBS (r) & 0.0753 $\pm$ 0.0031 &  2.56 $\pm$ 0.04 & 15.95 $\pm$ 1.13 & 0.0467\\
\hline 
\multicolumn{5}{|c|}{$M=30$, $\mu_0 = 0.0345$, $\sigma = 0.0503$}  \\ \hline
CCBS & 0.0740 $\pm$ 0.0029 &  5.41 $\pm$ 0.18 & 42 & 0.0339\\
CCBS (r) & 0.0740 $\pm$ 0.0029 &  5.62 $\pm$ 0.19 &  42& 0.0339\\
GCBS & 0.0737 $\pm$ 0.0029 &  3.24 $\pm$ 0.04 & 41 & 0.0344\\
GCBS (r) & 0.0742 $\pm$ 0.0030 &  2.74 $\pm$ 0.07 & 33.39 $\pm$ 1.43 & 0.0326\\
\hline
\end{tabular}
\end{center}
\end{table}

\begin{table}[ht]
\caption{RMSE. 100 runs, 2000 pxl., 8 endmembers (Pavia), SNR=21dB, GBM, SK-Hype. $\mu_0$ computed using Equation~\eqref{eq:mu_0} for a given $M$, and $\sigma$ is found solving problem~\eqref{eq:sigmaOpt}.}
\label{tab:PaviaSynthetic_GBM}
\begin{center}
\begin{tabular}{|l|c|c|c|c|}\hline
Strategy & RMSE $\pm$ STD &  Av. Time & $N_b$ & $\mu$ \\ \hline
SK-Hype & 0.0810 $\pm$ 0.0035 &  15.2468 $\pm$ 0.3231 & 103 & -\\
\hline
GKKM & 0.0852 $\pm$ 0.0038 &  5.69 $\pm$ 0.01 & 5 & 0.5347\\
\hline
\multicolumn{5}{|c|}{$M=5$, $\mu_0 = 0.25$, $\sigma = 0.2385$}  \\ \hline
CCBS & 0.0845 $\pm$ 0.0037 &  4.62 $\pm$ 0.05 & 6 & 0.2402\\
CCBS (rand) & 0.0845 $\pm$ 0.0037 &  4.64 $\pm$ 0.05 &  6& 0.2395\\
GCBS & 0.0848 $\pm$ 0.0037 &  4.54 $\pm$ 0.02 & 6 & 0.2338\\
GCBS (rand) & 0.0862 $\pm$ 0.0038 &  5.02 $\pm$ 0.21 & 4.89 $\pm$ 0.37 & 0.1812\\
\hline
\multicolumn{5}{|c|}{$M=10$, $\mu_0 = 0.1111$, $\sigma = 0.1$}  \\ \hline
CCBS & 0.0813 $\pm$ 0.0035 &  3.51 $\pm$ 0.04 & 12 & 0.1098\\
CCBS (rand) & 0.0813 $\pm$ 0.0035 &  3.53 $\pm$ 0.05 &  12 & 0.1098\\
GCBS & 0.0824 $\pm$ 0.0035 &  3.65 $\pm$ 0.03 & 12 & 0.1080\\
GCBS (rand) & 0.0832 $\pm$ 0.0036 &  3.76 $\pm$ 0.12 & 9.58 $\pm$ 0.75 & 0.0907\\
\hline 
\multicolumn{5}{|c|}{$M=20$, $\mu_0 = 0.0526$, $\sigma = 0.0498$}  \\ \hline
CCBS & 0.0795 $\pm$ 0.0034 &  3.43 $\pm$ 0.04 & 20 & 0.0383\\
CCBS (rand) & 0.0794 $\pm$ 0.0034 &  3.45 $\pm$ 0.04 &  20 & 0.0437\\
GCBS & 0.0795 $\pm$ 0.0034 &  3.49 $\pm$ 0.02 & 20 & 0.0499\\
GCBS (rand) & 0.0804 $\pm$ 0.0035 &  3.45 $\pm$ 0.07 & 16.55 $\pm$ 0.88 & 0.0408\\
\hline 
\multicolumn{5}{|c|}{$M=30$, $\mu_0 = 0.0345$, $\sigma = 0.0353$}  \\ \hline
CCBS & 0.0784 $\pm$ 0.0034 &  3.68 $\pm$ 0.03 & 25 & 0.0314\\
CCBS (rand) & 0.0784 $\pm$ 0.0033 &  3.68 $\pm$ 0.04 &  25 & 0.0311\\
GCBS & 0.0787 $\pm$ 0.0034 &  3.67 $\pm$ 0.03 & 25 & 0.0300\\
GCBS (rand) & 0.0790 $\pm$ 0.0034 &  3.54 $\pm$ 0.06 & 21.09 $\pm$ 1.02 & 0.0282\\
\hline
\end{tabular}
\end{center}
\end{table}

\begin{table}[ht]
\caption{RMSE. 100 runs, 2000 pxl., 8 endmembers (Pavia), SNR=21dB, PNMM, SK-Hype. $\mu_0$ computed using Equation~\eqref{eq:mu_0} for a given $M$, and $\sigma$ is found solving problem~\eqref{eq:sigmaOpt}.}
\label{tab:PaviaSynthetic_PNMM}
\begin{center}
\begin{tabular}{|l|c|c|c|c|}\hline
Strategy & RMSE $\pm$ STD &  Av. Time & $N_b$ & $\mu$ \\ \hline
SK-Hype & 0.0839 $\pm$ 0.0035 &  14.6747 $\pm$ 0.3073 & 103 & -\\
\hline
GKKM & 0.0878 $\pm$ 0.0038 &  5.31 $\pm$ 0.02 & 5 & 0.5347\\
\hline
\multicolumn{5}{|c|}{$M=5$, $\mu_0 = 0.25$, $\sigma = 0.2385$}  \\ \hline
CCBS & 0.0861 $\pm$ 0.0037 &  4.34 $\pm$ 0.04 & 6 & 0.2402\\
CCBS (r) & 0.0861 $\pm$ 0.0037 &  4.34 $\pm$ 0.05 &  6 & 0.2395\\
GCBS & 0.0877 $\pm$ 0.0038 &  4.17 $\pm$ 0.02 & 6 & 0.2338\\
GCBS (r) & 0.0882 $\pm$ 0.0039 &  4.56 $\pm$ 0.23 & 4.89 $\pm$ 0.37 & 0.1812\\
\hline
\multicolumn{5}{|c|}{$M=10$, $\mu_0 = 0.1111$, $\sigma = 0.1$}  \\ \hline
CCBS & 0.0835 $\pm$ 0.0035 &  3.27 $\pm$ 0.03 & 12 & 0.1098\\
CCBS (r) & 0.0835 $\pm$ 0.0035 &  3.25 $\pm$ 0.04 &  12 & 0.1098\\
GCBS & 0.0852 $\pm$ 0.0035 &  3.32 $\pm$ 0.01 & 12 & 0.1080\\
GCBS (r) & 0.0857 $\pm$ 0.0036 &  3.38 $\pm$ 0.08 & 9.58 $\pm$ 0.75 & 0.0907\\
\hline 
\multicolumn{5}{|c|}{$M=20$, $\mu_0 = 0.0526$, $\sigma = 0.0498$}  \\ \hline
CCBS & 0.0817 $\pm$ 0.0034 &  3.22 $\pm$ 0.04 & 20 & 0.0383\\
CCBS (r) & 0.0817 $\pm$ 0.0034 &  3.23 $\pm$ 0.05 &  20 & 0.0437\\
GCBS & 0.0817 $\pm$ 0.0034 &  3.27 $\pm$ 0.02 & 20 & 0.0499\\
GCBS (r) & 0.0828 $\pm$ 0.0035 &  3.24 $\pm$ 0.05 & 16.55 $\pm$ 0.88 & 0.0408\\
\hline 
\multicolumn{5}{|c|}{$M=30$, $\mu_0 = 0.0345$, $\sigma = 0.0353$}  \\ \hline
CCBS & 0.0804 $\pm$ 0.0033 &  3.43 $\pm$ 0.05 & 25 & 0.0314\\
CCBS (r) & 0.0803 $\pm$ 0.0033 &  3.45 $\pm$ 0.03 &  25 & 0.0311\\
GCBS & 0.0806 $\pm$ 0.0033 &  3.48 $\pm$ 0.05 & 25 & 0.0300\\
GCBS (r) & 0.0810 $\pm$ 0.0034 &  3.33 $\pm$ 0.06 & 21.09 $\pm$ 1.02 & 0.0282\\
\hline
\end{tabular}
\end{center}
\end{table}

\subsection{Simulation with real data}
When working with real data ground truth for the fractional abundances are rarely available. Thus, we compare the abundance estimation results obtained using a full band approach and using the proposed band selection strategy.  First, the data is unmixed using the SK-Hype algorithm using all the available spectral bands, what yields the estimated abundances $\balpha_n^{\text{skp}},\, n=1,\ldots,N$. The unmixing is then done for all each of the BS methods presented in Section~\ref{sec:BS}. Generically denominating the BS-based estimated abundances as $\balpha_n^{\text{bs}},\, n=1,\ldots,N$, the RMSE between the SK-Hype abundances and those obtained using a given BS algorithm is computed as 

\begin{equation}
\label{eq:RmseRealData}
\text{RMSE}=\sqrt{\sum_{n=1}^{N}\|\balpha^{\text{skp}}_n-\balpha^{\text{bs}}_n\|^2/(N\times R)}. 
\end{equation}

%\subsubsection{Pavia University}

% 
% \begin{figure}
% \captionsetup[subfigure]{position=b}
%          \centering        
%         %\begin{subfigure}[b]{0.21\textwidth}
%         \begin{subfigure}[b]{0.15\textwidth}
%                 \includegraphics[width=\textwidth]{../Figure/paviaU_Rec_5_30_50-crop}
%                 \caption{Pavia University representation.}
%                 \label{fig:paviaU}
%         \end{subfigure} %\vspace{0.5cm}
%          %\qquad %add desired spacing between images, e. g. ~, \quad, \qquad, \hfill etc.
%           %(or a blank line to force the subfigure onto a new line)       
%         %\begin{subfigure}[b]{0.263\textwidth}
%         \begin{subfigure}[b]{0.1873\textwidth}
%                 \includegraphics[width=\textwidth]{../Figure/paviaU_gt-crop}
%                 \caption{Ground truth for the Pavia University scene.}
%                 \label{fig:paviaU_gt}
%         \end{subfigure}
%         \caption{Pavia University. In (a) the Pavia University HI is represented using the bands 5, 30, and 50. In (b) the classified areas are labelled from 1 to 9, while 0 corresponds to unclassified areas.}\label{fig:pavia}
% \end{figure}

%\subsubsection{Cuprite}

\begin{figure}
 \centering
 \includegraphics[width=0.25\textwidth]{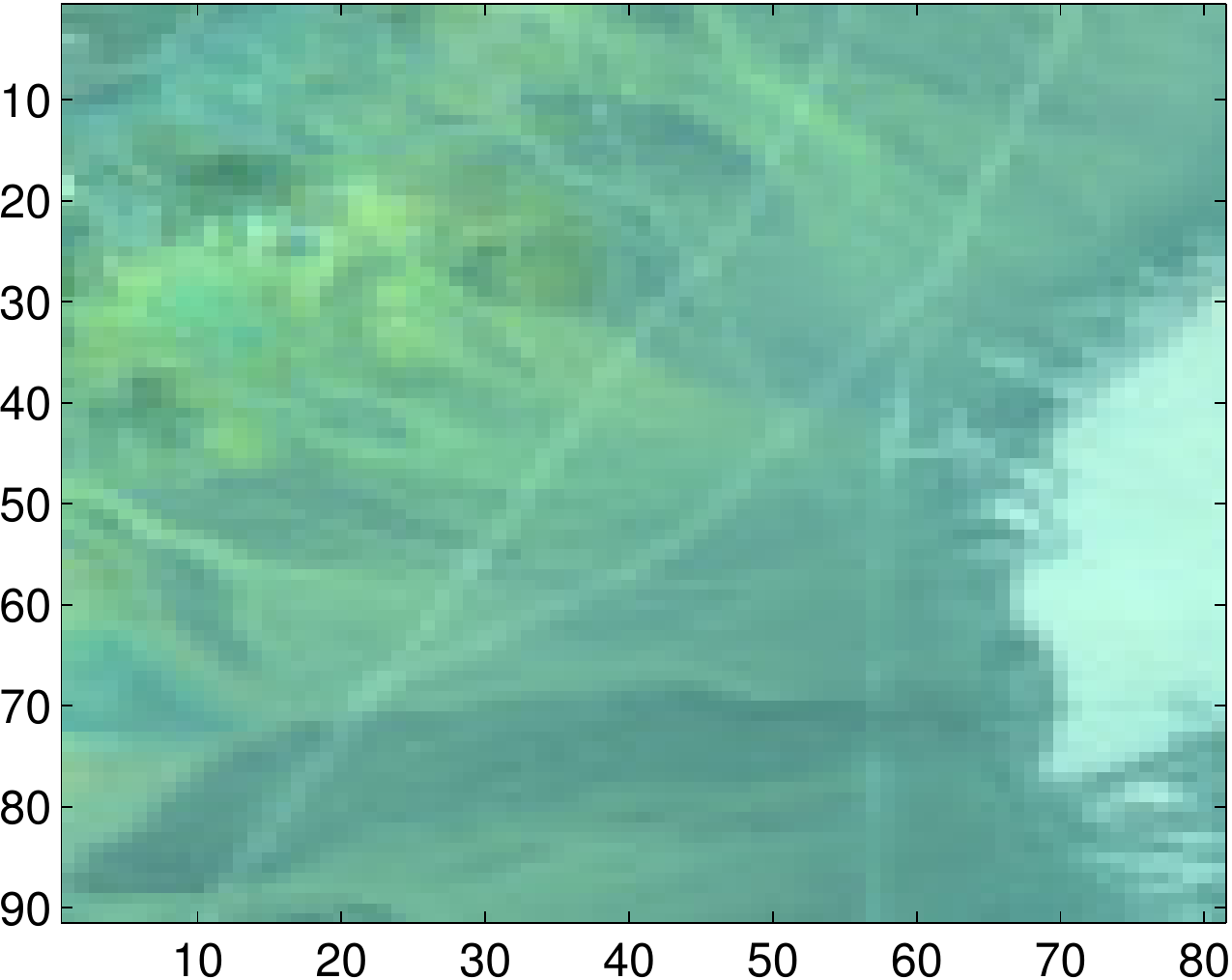}
 \caption{Cuprite scene used in~\cite{Imbiriba2016_tip}.}
 \label{fig:cupriteimg}
\end{figure}

The images used are shown in Figure~\ref{fig:cupriteimg} and Figure~\ref{fig:pavia}. The first image is a scene from the Cuprite mining field site in Nevada, acquired by the AVIRIS instrument. It has originally 224 spectral bands, from which we have removed the water absorption bands, resulting in 188 bands. This scene has 7371 pixels and previous analysis identified five minerals (Sphene, Montmorillonite, Kaolinite, Dumortierite, and Pyrope) to have strong components in this particular region~\cite{Imbiriba2016_tip}. The endmember matrix was extracted using the VCA algorithm~\cite{Nascimento2005}. The second image is the scene from the Pavia University described in Section~\ref{sec:SimSynthetic}. It has 207400 pixels and the endmembers were also extracted using VCA, see Section \ref{sec:SimSynthetic}.

Tables~\ref{tab:cupriteReal} and~\ref{tab:paviaReal} show the abundance RMSE results obtained using \eqref{eq:RmseRealData}. For both tables, the RMSE performance is compatible to that obtained using synthetic images, and the savings in computational complexity can be inferred from the CPU time reduction by a factor of at least 13 (for $M=30)$ for the Cuprite scene and at least 3 (for $M=30$) for the Pavia scene. In comparing CCBS and GCBS with GKKM one should note the significant reduction obtained in dictionary coherence for the same model complexity ($N_b$).

\begin{table}
\caption{Cuprite image. RMSE between the abundances estimated with SK-Hype (all bands) and BS + SK-Hype.}
\label{tab:cupriteReal}
\begin{center}
\begin{tabular}{|l|c|c|c|c|}\hline
Strategy & RMSE $\pm$ STD &  CPU Time & $N_b$ & $\mu$\\ \hline
SK-Hype & - & 282.42  & 188 & - \\
\hline
GKKM & 0.0777 $\pm$ 0.0036 & 19.289 & 13 & 0.8162\\ \hline
\multicolumn{5}{|c|}{$M=5$, $\mu_0 = 0.25$, $\sigma = 0.0963$}  \\ \hline
CCBS & 0.0805 $\pm$ 0.0038 & 18.4835 & 9 & 0.2495\\ 
GCBS & 0.0833 $\pm$ 0.0040 & 17.7114 & 9 & 0.2483\\ 
\hline
\multicolumn{5}{|c|}{$M=10$, $\mu_0 = 0.1111$, $\sigma = 0.0489$}  \\ \hline
CCBS & 0.0659 $\pm$ 0.0027 & 15.2023 & 16 & 0.1090\\ 
GCBS & 0.0695 $\pm$ 0.0029 & 14.5721 & 15 & 0.1090\\ 
\hline 
\multicolumn{5}{|c|}{$M=20$, $\mu_0 = 0.0526$, $\sigma = 0.0260$}  \\ \hline
CCBS & 0.0477 $\pm$ 0.0015 & 17.0942 & 25 & 0.0471\\ 
GCBS & 0.0484 $\pm$ 0.0015 & 16.9595 & 25 & 0.0493\\ 
\hline 
\multicolumn{5}{|c|}{$M=30$, $\mu_0 = 0.0345$, $\sigma = 0.0178$}  \\ \hline
CCBS & 0.0378 $\pm$ 0.0010 & 20.6932 & 35 & 0.0333\\ 
GCBS & 0.0395 $\pm$ 0.0011 & 20.4790 & 34 & 0.0300\\ 
\hline
\end{tabular}
\end{center}
\end{table}

% \begin{table}
% \caption{Cuprite image. RMSE between the abundances estimated with SK-Hype (all bands) and BS + SK-Hype.}
% \label{tab:cupriteReal}
% \begin{center}
% \begin{tabular}{|l|c|c|c|c|}\hline
% Strategy & RMSE $\pm$ STD &  CPU Time & $N_b$ & $\mu$\\ \hline
% SK-Hype & - & 356.44  & 188 & - \\
% \hline
% \multicolumn{5}{|c|}{$M=5$, $\mu_0 = 0.25$, $\sigma = 0.0963$}  \\ \hline
% CCBS & 0.0805 $\pm$ 0.0038 & 19.6558 & 9 & 0.2495\\ 
% GCBS & 0.0833 $\pm$ 0.0040 & 19.0268 & 9 & 0.2483\\ 
% GKKM & 0.0790 $\pm$ 0.0038 & 19.1646 & 9 & 0.7945\\
% \hline
% \multicolumn{5}{|c|}{$M=10$, $\mu_0 = 0.1111$, $\sigma = 0.0489$}  \\ \hline
% CCBS & 0.0659 $\pm$ 0.0027 & 16.2444 & 16 & 0.1090\\ 
% GCBS & 0.0695 $\pm$ 0.0029 & 15.6020 & 15 & 0.1090\\ 
% GKKM & 0.0692 $\pm$ 0.0029 & 16.1324 & 16 & 0.5224\\ 
% \hline 
% \multicolumn{5}{|c|}{$M=20$, $\mu_0 = 0.0526$, $\sigma = 0.0260$}  \\ \hline
% CCBS & 0.0477 $\pm$ 0.0015 & 18.2195 & 25 & 0.0471\\ 
% GCBS & 0.0484 $\pm$ 0.0015 & 18.0577 & 25 & 0.0493\\ 
% GKKM & 0.0519 $\pm$ 0.0018 & 18.2642 & 25 & 0.5324\\ 
% \hline 
% \multicolumn{5}{|c|}{$M=30$, $\mu_0 = 0.0345$, $\sigma = 0.0178$}  \\ \hline
% CCBS & 0.0378 $\pm$ 0.0010 & 21.6524 & 35 & 0.0333\\ 
% GCBS & 0.0395 $\pm$ 0.0011 & 21.4219 & 34 & 0.0300\\ 
% GKKM & 0.0386 $\pm$ 0.0010 & 22.3098 & 35 & 0.5446\\ 
% \hline
% \end{tabular}
% \end{center}
% \end{table}

\begin{table}
\caption{Pavia University image. RMSE between the abundances estimated with SK-Hype (all bands) and BS + SK-Hype.}
\label{tab:paviaReal}
\begin{center}
\begin{tabular}{|l|c|c|c|c|}\hline
Strategy & RMSE $\pm$ STD &  CPU Time & $N_b$ & $\mu$\\ \hline
SK-Hype & - & 1740.47 & 103 & -\\
\hline
GKKM & 0.0446 $\pm$ 0.0015 & 568.10 & 13 & 0.5066\\
\hline
\multicolumn{5}{|c|}{$M=5$, $\mu_0 = 0.25$, $\sigma = 0.2492$}  \\ \hline
CCBS & 0.0659 $\pm$ 0.0037 & 513.21 & 6 & 0.2499\\ 
GCBS & 0.0650 $\pm$ 0.0036 & 533.48 & 6 & 0.2499\\ 
\hline
\multicolumn{5}{|c|}{$M=10$, $\mu_0 = 0.1111$, $\sigma = 0.1017$}  \\ \hline
CCBS & 0.0435 $\pm$ 0.0016 & 495.13 & 12 & 0.1024\\ 
GCBS & 0.0500 $\pm$ 0.0023 & 497.92 & 12 & 0.1019\\ 
\hline 
\multicolumn{5}{|c|}{$M=20$, $\mu_0 = 0.0526$, $\sigma = 0.0503$}  \\ \hline
CCBS & 0.0301 $\pm$ 0.0008 & 488.67 & 21 & 0.0433\\ 
GCBS & 0.0309 $\pm$ 0.0009 & 488.66 & 21 & 0.0472\\
\hline 
\multicolumn{5}{|c|}{$M=30$, $\mu_0 = 0.0345$, $\sigma = 0.0336$}  \\ \hline
CCBS & 0.0260 $\pm$ 0.0007 & 535.64 & 26 & 0.0336\\ 
GCBS & 0.0263 $\pm$ 0.0007 & 538.63 & 26 & 0.0336\\ 
\hline
\end{tabular}
\end{center}
\end{table}

\subsubsection{Reconstruction error (Cuprite)}
One way to try to compare results with real images would be the reconstruction error. In Table~\ref{tab:recErrorCuprite} the results for the reconstruction error for the Cuprite scene are summarized. In this simulation the value of $M$ was increased up to $M=2000$ to produce larger dictionaries as examine the behaviour of the reconstruction error as the number of selected bands $N_b$ increases. However, even using $M=2000$ was not enough to use all 188 bands. This is expected since the maximal cardinality of the dictinary is bounded, see~\cite{Richard2009}.

\begin{table}[h]
\caption{Reconstruction Error for the Cuprite Scene.}\label{tab:recErrorCuprite}
\centering
\begin{tabular}{|l|c|c|c|c|}
\hline
Strategy & RMSE $\pm$ STD &  Time & $N_b$ & $\mu$\\\hline
SK-Hype & 0.0006 $\pm$ 0.0000 &  184.2852 & 188 & - \\\hline
GKKM & 0.0064 $\pm$ 0.0000 & 17.0144 & 13 & 0.7982\\ \hline
\multicolumn{5}{|c|}{$M = 5$, $\mu_0 = 0.2500$, $\sigma = 0.0916$}\\ \hline
CCBS & 0.0155 $\pm$ 0.0002 & 13.5109 & 9 & 0.2454\\ 
GCBS & 0.0129 $\pm$ 0.0001 & 15.9691 & 9 & 0.2454\\\hline
\multicolumn{5}{|c|}{$M = 30$, $\mu_0 = 0.0345$, $\sigma = 0.0174$}\\\hline
CCBS & 0.0101 $\pm$ 0.0001 & 26.1530 & 36 & 0.0336\\ 
GCBS & 0.0102 $\pm$ 0.0001 & 26.1337 & 35 & 0.0341\\ \hline
\multicolumn{5}{|c|}{$M = 50$, $\mu_0 = 0.0204$, $\sigma = 0.0113$}\\\hline
CCBS & 0.0099 $\pm$ 0.0001 & 38.1907 & 49 & 0.0199\\ 
GCBS & 0.0092 $\pm$ 0.0001 & 19.4064 & 49 & 0.0202\\ \hline
\multicolumn{5}{|c|}{$M = 70$, $\mu_0 = 0.0145$, $\sigma = 0.0087$}\\\hline
CCBS & 0.0089 $\pm$ 0.0001 & 24.5304 & 61 & 0.0141\\ 
GCBS & 0.0089 $\pm$ 0.0001 & 26.5687 & 61 & 0.0130\\ \hline
\multicolumn{5}{|c|}{$M = 120$, $\mu_0 = 0.0084$, $\sigma = 0.0059$}\\\hline
CCBS & 0.0079 $\pm$ 0.0000 & 42.4282 & 84 & 0.0084\\ 
GCBS & 0.0077 $\pm$ 0.0000 & 62.1598 & 84 & 0.0084\\ \hline
\multicolumn{5}{|c|}{$M = 150$, $\mu_0 = 0.0067$, $\sigma = 0.0051$}\\\hline
CCBS & 0.0074 $\pm$ 0.0000 & 55.4628 & 93 & 0.0067\\ 
GCBS & 0.0076 $\pm$ 0.0000 & 56.0754 & 92 & 0.0067\\ \hline
\multicolumn{5}{|c|}{$M = 188$, $\mu_0 = 0.0053$, $\sigma = 0.0044$}\\\hline
CCBS & 0.0076 $\pm$ 0.0000 & 54.0429 & 98 & 0.0043\\ 
GCBS & 0.0075 $\pm$ 0.0000 & 56.0656 & 97 & 0.0047\\ \hline
\multicolumn{5}{|c|}{$M = 500$, $\mu_0 = 0.0020$, $\sigma = 0.0026$}\\\hline
CCBS & 0.0067 $\pm$ 0.0000 & 95.6673 & 122 & 0.0017\\ 
GCBS & 0.0068 $\pm$ 0.0000 & 93.1751 & 122 & 0.0016\\ \hline
\multicolumn{5}{|c|}{$M = 1000$, $\mu_0 = 0.0010$, $\sigma = 0.0019$}\\\hline
CCBS & 0.0067 $\pm$ 0.0000 & 87.0426 & 131 & 0.0009\\ 
GCBS & 0.0067 $\pm$ 0.0000 & 96.7277 & 131 & 0.0009\\ \hline
\multicolumn{5}{|c|}{$M = 2000$, $\mu_0 = 0.0005$, $\sigma = 0.0015$}\\\hline
CCBS & 0.0063 $\pm$ 0.0000 & 107.6217 & 137 & 0.0004\\ 
GCBS & 0.0063 $\pm$ 0.0000 & 98.2356 & 137 & 0.0004\\ \hline
\end{tabular}
\end{table}

\subsection{{RELAB} data}
The RELAB data considered in~\cite{mustard1987quantitative, mustard1989photometric} has laboratory measured reflectances, and thus provides ground truth. The data consists of intimate mixtures of minerals (Anorthite, Olivine, Enstatite, and Magnetite) that were crushed and mixed together. The data is composed by the reflectances of the 4 pure minerals (endmembers) and of binary (Olivine/Enstatite, Olivine/Magnetite, and Olivine/Anorthite) and ternary (Olivine/Anorthite/Enstatite) mixtures. Each binary combination of minerals has 5 mixtures with different abundances for each endmembers (ranging form 0.1 to 0.95). The ternary mineral combinations have 7 spectra,  considering also different abundances. These spectra could be properly located in the RELAB dataset thanks to the help of Prof. John F. Mustard. 

We performed simulations following the procedure described in Section~\ref{sec:SimSynthetic}. The obtained results are summarized in the Tables 1 to 5 below, where the good performance of the proposed BS methods can be verified. We note that the proposed BS algorithms produced results that are close to the ones obtained using the full band SK-Hype algorithm. The best result using a BS strategy were obtained by the CCBS algorithm, which also produced the smallest RMSE when a ternary mixture was considered (see Table~\ref{tab:TernaryMix}).

Tables~\ref{tab:relab1} to~\ref{tab:relab4} present simulations using mixtures of two endmembers. 
In these tables the full band SK-Hype algorithm presented the smallest RMSE for the abundance estimations. Although the full band SK-Hype presents the smallest RMSE, the RMSE obtained using the proposed BS methods (CCBS and GCBS) are comparable, specially for $M=30$, indicating the possibility of a significant reduction in computational complexity. 

The tables show GKKM RMSE results that are worse than those using the proposed methods in three out of four cases (tables~\ref{tab:relab1}, \ref{tab:relab3} and \ref{tab:relab4}), for similar number of bands. Table~\ref{tab:relab2} shows slightly better results for GKKM for the mixture Olivine/Magnetite. 

Please note that these results are based on averages of five realizations only and, therefore, their statistical significance has to be taken with care.  This is the main reason why we have not included such results in the final manuscript, only showing them in this technical report.  These results provide some confidence that the results reported in the paper indicate the true potential of the proposed methods, but they can hardly be quoted as good performance evaluations in a comparative study among different techniques de to their low statistical significance.

Regarding the CPU Time elapsed by the algorithms, again we note the considerable effect of dealing with a reduced amount of data. Looking at the numbers in tables~\ref{tab:relab1}, \ref{tab:relab3} and \ref{tab:relab4}, one notices that the greedy approach GCBS needs CPU Times that are 15 to 20 times smaller than the full band solution for $M = 30$. This CPU time reduction is even larger for $M<30$. One notices, however, a large variation in CPU times for the MCP based algorithm (CCBS), especially for $M=30$. Tables~\ref{tab:relab1} and~\ref{tab:relab2} show CCBS CPU times ($M=30$) that are even larger then the CPU times for the full band solution. The noticeable differences for different data sets are due to the solution of the maximal clique problem. Since MCP are NP-hard problems, the required CPU time for its solution can significantly change for different data sets. The fact that binary RELAB mixtures considered here are composed of only small numbers of mixtures (5 pixels) makes the processing time of solving a MCP more evident. This required MCP time is greatly diluted when larger data sets are considered, as could be verified in the results presented in the paper. This indicates that the CCBS is more advantageous for larger datasets, a common situation in hyperspectral image processing. The GKKM algorithm has considerable CPU Time. Although the parameter and band selection procedure consumes a great amount of time, what should also be diluted for bigger datasets. However, the simulations with bigger datasets (synthetic and real) also indicate greater CPU Times required by the GKKM, when compared with the proposed algorithms.

The results shown in Table~\ref{tab:TernaryMix} for the ternary mixture lead to similar conclusions regarding the RMSE and CPU time required by the algorithms. However, applying the proposed methods resulted in improvements in the RMSE results for $M>10$ when compared with the full band SK-Hype. The RMSE obtained with the GKKM is comparable with the full band SK-Hype. Nevertheless, the above comments on the statistical significance of the result apply.

\begin{table}[ht]
\centering
\caption{Olivine/Enstatite}\label{tab:relab1}
\begin{tabular}{|c|c|c|c|c|}
\hline
%\multicolumn{5}{|c|}{Olivine/Enstatite}\\ \hline
Strategy & RMSE $\pm$ STD &  Time & $N_b$ & $\mu$\\\hline
SK-Hype & 0.0442 $\pm$ 0.0011 &  0.2499 & 211 & -\\\hline
GKKM & 0.1883 $\pm$ 0.0317 & 1.8460 & 14 & 0.7969\\\hline 
\multicolumn{5}{|c|}{$M = 5$, $\mu_0 = 0.2500$, $\sigma = 0.1034$}\\\hline
CCBS & 0.2045 $\pm$ 0.0375 & 0.1001 & 7 & 0.2490\\ 
GCBS & 0.2441 $\pm$ 0.0500 & 0.0374 & 5 & 0.2097\\ \hline
\multicolumn{5}{|c|}{$M = 10$, $\mu_0 = 0.1111$, $\sigma = 0.0524$}\\\hline
CCBS & 0.1430 $\pm$ 0.0192 & 0.0702 & 14 & 0.1079\\ 
GCBS & 0.1505 $\pm$ 0.0209 & 0.0191 & 13 & 0.1078\\ \hline
\multicolumn{5}{|c|}{$M = 20$, $\mu_0 = 0.0526$, $\sigma = 0.0273$}\\\hline
CCBS & 0.0907 $\pm$ 0.0069 & 0.2874 & 24 & 0.0517\\ 
GCBS & 0.0941 $\pm$ 0.0081 & 0.0245 & 21 & 0.0492\\ \hline
\multicolumn{5}{|c|}{$M = 30$, $\mu_0 = 0.0345$, $\sigma = 0.0182$}\\\hline
CCBS & 0.0705 $\pm$ 0.0035 & 0.6403 & 34 & 0.0339\\ 
GCBS & 0.0674 $\pm$ 0.0034 & 0.0162 & 32 & 0.0331\\ \hline
\end{tabular}
\end{table}

\begin{table}[ht]
\centering
\caption{Olivine/Magnetite}\label{tab:relab2}
\begin{tabular}{|c|c|c|c|c|}
\hline
%\multicolumn{5}{|c|}{Olivine/Magnetite}\\\hline
Strategy & RMSE $\pm$ STD &  Time & $N_b$ & $\mu$\\\hline
SK-Hype & 0.3279 $\pm$ 0.1596 &  0.2145 & 211 & -\\\hline
GKKM & 0.3269 $\pm$ 0.1594 & 0.6759 & 7 & 0.9220\\ \hline
\multicolumn{5}{|c|}{$M = 5$, $\mu_0 = 0.2500$, $\sigma = 0.0450$}\\\hline
CCBS & 0.3427 $\pm$ 0.1751 & 0.0606 & 6 & 0.2461\\ 
GCBS & 0.3446 $\pm$ 0.1767 & 0.0209 & 6 & 0.2439\\ \hline
\multicolumn{5}{|c|}{$M = 10$, $\mu_0 = 0.1111$, $\sigma = 0.0201$}\\\hline
CCBS & 0.3361 $\pm$ 0.1686 & 0.0671 & 12 & 0.1101\\ 
GCBS & 0.3428 $\pm$ 0.1745 & 0.0142 & 9 & 0.0840\\ \hline
\multicolumn{5}{|c|}{$M = 20$, $\mu_0 = 0.0526$, $\sigma = 0.0104$}\\\hline
CCBS & 0.3327 $\pm$ 0.1647 & 0.3483 & 21 & 0.0514\\ 
GCBS & 0.3361 $\pm$ 0.1680 & 0.0106 & 19 & 0.0518\\ \hline
\multicolumn{5}{|c|}{$M = 30$, $\mu_0 = 0.0345$, $\sigma = 0.0071$}\\ \hline
CCBS & 0.3318 $\pm$ 0.1637 & 3.6450 & 28 & 0.0331\\ 
GCBS & 0.3354 $\pm$ 0.1669 & 0.0132 & 23 & 0.0336\\ \hline
\end{tabular}
\end{table}

\begin{table}[ht]
\centering
\caption{Olivine/Anorthite}\label{tab:relab3}
\begin{tabular}{|c|c|c|c|c|}
\hline
%\multicolumn{5}{|c|}{Olivine/Anorthite}\\\hline
Strategy & RMSE $\pm$ STD &  Time & $N_b$ & $\mu$\\\hline
SK-Hype & 0.1249 $\pm$ 0.0131 &  0.3467 & 211 & -\\ \hline
GKKM & 0.2322 $\pm$ 0.0508 & 0.7728 & 8 & 0.8841\\ \hline
\multicolumn{5}{|c|}{$M = 5$, $\mu_0 = 0.2500$, $\sigma = 0.0538$}\\\hline
CCBS & 0.2112 $\pm$ 0.0426 & 0.0690 & 8 & 0.2396\\ 
GCBS & 0.2251 $\pm$ 0.0491 & 0.0237 & 6 & 0.2357\\ \hline
\multicolumn{5}{|c|}{$M = 10$, $\mu_0 = 0.1111$, $\sigma = 0.0276$}\\\hline
CCBS & 0.1802 $\pm$ 0.0290 & 0.0682 & 14 & 0.1095\\ 
GCBS & 0.1839 $\pm$ 0.0304 & 0.0176 & 13 & 0.1104\\ \hline
\multicolumn{5}{|c|}{$M = 20$, $\mu_0 = 0.0526$, $\sigma = 0.0155$}\\ \hline
CCBS & 0.1534 $\pm$ 0.0150 & 0.0851 & 23 & 0.0521\\ 
GCBS & 0.1655 $\pm$ 0.0198 & 0.0195 & 20 & 0.0507\\ \hline
\multicolumn{5}{|c|}{$M = 30$, $\mu_0 = 0.0345$, $\sigma = 0.0111$}\\ \hline
CCBS & 0.1399 $\pm$ 0.0105 & 0.0866 & 30 & 0.0325\\ 
GCBS & 0.1426 $\pm$ 0.0111 & 0.0167 & 27 & 0.0296\\ \hline
\end{tabular}
\end{table}

\begin{table}[ht]
\centering
\caption{Anorthite/Enstatite}\label{tab:relab4}
\begin{tabular}{|c|c|c|c|c|}
\hline
%\multicolumn{5}{|c|}{Anorthite/Enstatite}\\ \hline
Strategy & RMSE $\pm$ STD &  Time & $N_b$ & $\mu$\\\hline
SK-Hype & 0.1900 $\pm$ 0.0303 &  0.3260 & 211 & -\\\hline
GKKM & 0.2577 $\pm$ 0.0590 & 0.7722 & 9 & 0.8765\\ \hline
\multicolumn{5}{|c|}{$M = 5$, $\mu_0 = 0.2500$, $\sigma = 0.0606$}\\\hline
CCBS & 0.2454 $\pm$ 0.0498 & 0.0594 & 8 & 0.2486\\ 
GCBS & 0.2503 $\pm$ 0.0568 & 0.0187 & 7 & 0.2435\\ \hline
\multicolumn{5}{|c|}{$M = 10$, $\mu_0 = 0.1111$, $\sigma = 0.0302$}\\\hline
CCBS & 0.2297 $\pm$ 0.0406 & 0.0678 & 12 & 0.1106\\ 
GCBS & 0.2478 $\pm$ 0.0504 & 0.0159 & 10 & 0.0995\\ \hline
\multicolumn{5}{|c|}{$M = 20$, $\mu_0 = 0.0526$, $\sigma = 0.0167$}\\\hline
CCBS & 0.1961 $\pm$ 0.0244 & 0.0809 & 22 & 0.0517\\ 
GCBS & 0.2126 $\pm$ 0.0291 & 0.0182 & 18 & 0.0493\\ \hline
\multicolumn{5}{|c|}{$M = 30$, $\mu_0 = 0.0345$, $\sigma = 0.0119$}\\\hline
CCBS & 0.1938 $\pm$ 0.0265 & 0.0823 & 30 & 0.0342\\ 
GCBS & 0.1988 $\pm$ 0.0263 & 0.0213 & 28 & 0.0343\\ \hline
\end{tabular}
\end{table}

\begin{table}[ht]
\centering
\caption{Olivine/Arnothite/Enstatite}\label{tab:TernaryMix}
\begin{tabular}{|c|c|c|c|c|}
\hline
%\multicolumn{5}{|c|}{Olivine/Arnothite/Enstatite}\\\hline
Strategy & RMSE $\pm$ STD &  Time & $N_b$ & $\mu$\\\hline
SK-Hype & 0.1320 $\pm$ 0.0161 &  0.3034 & 211 & -\\\hline
GKKM & 0.1325 $\pm$ 0.0150 & 1.8930 & 15 & 0.7694\\ \hline
\multicolumn{5}{|c|}{$M = 5$, $\mu_0 = 0.2500$, $\sigma = 0.1084$}\\\hline
CCBS & 0.1358 $\pm$ 0.0154 & 0.0550 & 7 & 0.2441\\ 
GCBS & 0.1585 $\pm$ 0.0229 & 0.0227 & 5 & 0.2212\\ \hline
\multicolumn{5}{|c|}{$M = 10$, $\mu_0 = 0.1111$, $\sigma = 0.0589$}\\\hline
CCBS & 0.1224 $\pm$ 0.0125 & 0.0740 & 14 & 0.1110\\ 
GCBS & 0.1323 $\pm$ 0.0158 & 0.0172 & 12 & 0.1076\\ \hline
\multicolumn{5}{|c|}{$M = 20$, $\mu_0 = 0.0526$, $\sigma = 0.0341$}\\\hline
CCBS & 0.1132 $\pm$ 0.0118 & 0.0731 & 22 & 0.0506\\ 
GCBS & 0.1179 $\pm$ 0.0131 & 0.0204 & 20 & 0.0491\\ \hline
\multicolumn{5}{|c|}{$M = 30$, $\mu_0 = 0.0345$, $\sigma = 0.0242$}\\\hline
CCBS & 0.1123 $\pm$ 0.0131 & 0.1679 & 29 & 0.0339\\ 
GCBS & 0.1166 $\pm$ 0.0143 & 0.0248 & 28 & 0.0323\\ \hline
\end{tabular}
\end{table}

\section{Conclusions}

In this paper we have proposed a centralized method for nonlinear unmixing of hyperspectral images, which employs band selection in in the reproducing kernel Hilbert space (RKHS). The proposed method is based on the coherence criterion, which incorporates a measure of the quality of the dictionary in the RKHS for the nonlinear unmixing. We have shown that the proposed BS approach is equivalent to solving a maximum clique problem (MCP). Contrary to competing methods that do not include an efficient choice of the model parameters, the proposed method requires only an initial guess on the number of selected bands. Simulation results employing both synthetic and real data illustrate the quality of the unmixing results obtained with the proposed method, which leads to abundance estimations as accurate as those obtained using the full-band SK-Hype method, at a small fraction of the computational cost.

\bibliographystyle{IEEEbib}
\bibliography{hyperspectral}

\end{document}